\documentclass[10pt,twocolumn,letterpaper]{article}

\usepackage[pagenumbers]{cvpr}

\usepackage{cuted} 

\usepackage{currfile} 

\usepackage{caption} 

\definecolor{cvprblue}{rgb}{0.21,0.49,0.74}
\usepackage[pagebackref,breaklinks,colorlinks,allcolors=cvprblue,hypertexnames=false]{hyperref}
\usepackage{multirow}
\usepackage{tabularx}

\def\confName{CVPR}
\def\confYear{2027}

\title{Embodied AI Fails the Mirror Test: How a Simple Mirror Breaks Robot Policies}

\author{%
Xuanchen Liu\textsuperscript{*}, Hejun Zhang\textsuperscript{*},
Boxuan Dong, Ying Wu,\\
Heshuai Li, Wen Gao, Yonghong Tian, Weihao Yu\textsuperscript{\dag}\\[2pt]
Peking University\\
{\small Code: \url{https://github.com/iPPLab/robodojo-mirror}}\\
{\small \textsuperscript{*}Equal contribution.\quad
\textsuperscript{\dag}Corresponding author:
\href{mailto:weihao@pku.edu.cn}{\texttt{weihao@pku.edu.cn}}}
}

\begin{document}

\maketitle

\begin{strip}
\centering
\includegraphics[width=\linewidth]{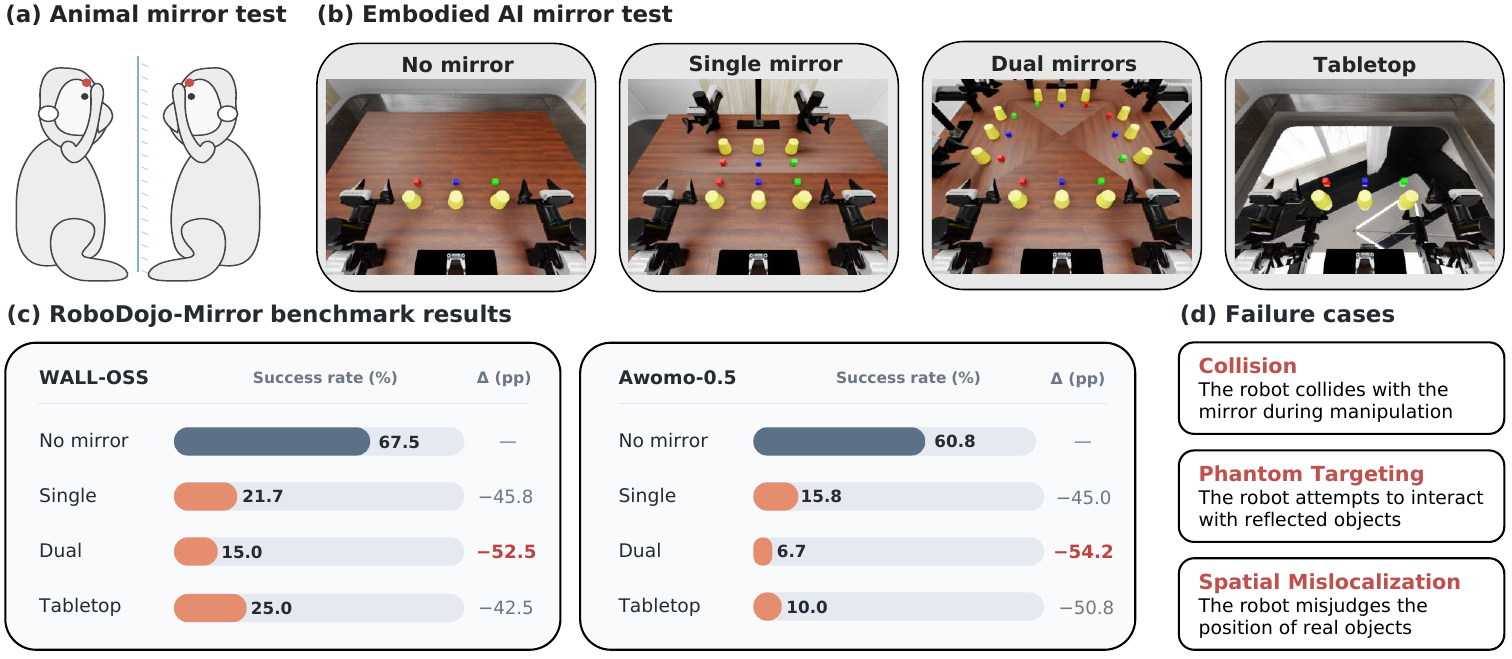}
\captionsetup{hypcap=false}
\captionof{figure}{\textbf{Overview of the mirror benchmark.}
(a) The animal mirror test motivates studying how robots act in the presence of reflections.
(b) Four experimental configurations: no mirror, a single mirror, dual mirrors, and a mirrored tabletop.
(c) Success rates of WALL-OSS and Awomo-0.5 across configurations, with changes relative to the no-mirror baseline in percentage points. Performance drops are computed from unrounded success rates.
(d) Three failure categories in mirror scenes: 
collision, involving unintended contact with the mirror surface; phantom targeting, where the robot acts toward a reflected object; and spatial mislocalization, where the robot incorrectly estimates a physical target's location.
}
\captionsetup{hypcap=true}
\label{fig:currfilebase}%
\end{strip}

\begin{abstract}
Can a simple mirror expose the limitations of today's robotic manipulation policies?
Inspired by the mirror self-recognition test in animal cognition, we investigate whether robot policies can reliably manipulate physical objects in scenes containing their reflections.
We introduce RoboDojo-Mirror, a benchmark built on RoboDojo that preserves task objectives, success criteria, and policy checkpoints while comparing a mirror-free baseline with single-mirror, dual-mirror, and mirrored-tabletop conditions.
Across five policies spanning Vision-Language-Action models, World Action Models, and agentic policies, every mirror condition reduces aggregate success rates: averaged over the three conditions, the drop ranges from 10.8 to 56.7 percentage points across policies, and it reaches 62.5 points on the mirrored tabletop.
Because a mirror is also a physical obstacle, we compare it against geometrically matched non-reflective panels; 
50--88\% of mirror-specific failures occur without recorded robot--panel contact, so direct collisions alone do not explain the degradation.
Trajectory and contact analyses instead reveal recurring behaviors consistent with phantom targeting of reflected objects and spatial mislocalization of real ones. Targeted diagnostics further characterize grasping near reflections, policy-dependent pushing progress, and physical-target interaction under main-camera occlusion.
Real-robot experiments with $\pi_{0.5}$ and FastWAM also show lower overall success rates under the upright-mirror and mirrored-tabletop conditions.
These findings reveal a shared vulnerability to reflective environments across diverse robot policies and motivate reflection-aware evaluation of embodied intelligence.
\end{abstract}

\section{Introduction}
\label{sec:intro}

Can an ordinary mirror reveal limitations in how robots connect visual observations to physical actions?
The classical mirror self-recognition test examines whether animals can use reflections to guide actions toward their own bodies~\cite{gallup1970selfrecognition}.
Inspired by this behavioral probe, we ask whether robot policies can reliably locate and manipulate physical objects when their observations also contain reflections of those objects.

Mirrors create a structured discrepancy between visual
appearance and actionable space. 
Reflections preserve
recognizable features and change with physical objects,
yet their apparent locations contain no additional manipulable targets. 
They can also adjoin real objects in images, forming composite contours, or reveal otherwise occluded targets. 
These properties enable tests of two complementary capabilities: maintaining reliable manipulation in the presence of reflections, and using reflected observations to act on occluded targets.

Recent advances in generalist robot policies have expanded the range of manipulation tasks that can be performed from visual and language inputs, yet whether these policies possess the two capabilities above remains unclear. 
Vision-Language-Action models directly generate actions from multimodal observations, while World Action Models jointly model actions and future observations~\cite{BlackK-RSS-25,ye2026worldactionmodelszeroshot}. 
Agentic policies provide another approach by selecting and invoking robot tools through repeated observation and reasoning~\cite{jia2026agentpolicyroboticmanipulation}. 
However, strong performance under standard evaluation conditions does not establish robustness to changes in visual evidence. Controlled perturbation benchmarks have demonstrated substantial performance degradation under variations in viewpoints, lighting, backgrounds, and other conditions~\cite{fei2026liberoplus}. 
Reflections are not covered by these perturbations: rather than altering appearance, they duplicate task-relevant content at locations with a known geometric relationship to the physical scene.

Existing mirror-based evaluations have begun to examine embodied agents’ self-centric perception and interaction, including tasks inspired by mirror self-recognition~\cite{guo2026mirrorbench}. We investigate a different operational setting: generalist policies performing external-object manipulation under controlled mirror configurations. The question posed above breaks down into three parts: how reflections affect task completion, whether observed degradation extends beyond the effects of an inserted physical panel, and what execution behaviors accompany these changes.

We introduce RoboDojo-Mirror, an evaluation benchmark built on RoboDojo~\cite{chen2026robodojo}. 
As illustrated in Fig.~\ref{fig:currfilebase}, it preserves the original manipulation objectives and success criteria while evaluating policies under four conditions: no mirror, a single mirror, dual mirrors, and a mirrored tabletop. 
On a shared set of eight tasks, we pair layouts and policy seeds across conditions to measure changes relative to each policy’s own mirror-free baseline. 
For selected policies, geometrically matched non-reflective panels provide additional controls for the effects of inserting physical structures. 
We further design three diagnostic scenarios, namely grasping on a mirrored tabletop, pushing an object connected to its reflection, and acting with reflections as the main visual cue, combining task outcomes with trajectory and contact observations.
This protocol enables researchers to assess reflection
robustness using existing policies and task evaluators,
and to characterize failures beyond aggregate success rates.

Across five policies spanning Vision-Language-Action
models, World Action Models, and agentic policies,
all three mirror conditions reduce aggregate success rates,
with average drops of 10.8--56.7 percentage points
across policies.
Matched panel comparisons show that 50--88\% of
mirror-specific failures occur without recorded
robot--panel contact, so direct collisions alone do
not explain the degradation.
The accompanying behaviors are consistent with phantom
targeting of reflected objects and spatial
mislocalization of real ones.
The three diagnostics further show more frequent empty
grasps, pushing progress that varies with policy and
contour, and limited
physical-target contact even with mirror-related hints.
Real-robot experiments with $\pi_{0.5}$
and FastWAM show a similar overall degradation trend.
Together, these results reveal a shared vulnerability
to reflections across diverse robot policies.
\section{Related Work}
\label{sec:related_work}

\subsection{Robotic Manipulation Benchmarks and Visual Robustness}

Robotic manipulation benchmarks provide complementary tools for
evaluating policy capabilities.
LIBERO studies knowledge transfer in lifelong robot
learning~\cite{liu2023libero}.
RoboTwin 2.0 combines scalable data generation with structured
domain randomization for bimanual
manipulation~\cite{chen2026robotwin2}, while RoboDojo provides
a unified simulation and real-world evaluation framework for
generalist manipulation policies~\cite{chen2026robodojo}.
Recent benchmarks place greater emphasis on robustness beyond
nominal task success.
LIBERO-Plus introduces controlled perturbations to object layouts,
camera viewpoints, robot initial states, language, lighting,
backgrounds, and sensor observations~\cite{fei2026liberoplus}.
LIBERO-VPro evaluates closed-loop visual robustness by perturbing
the evidence available during execution, including observation
degradation, camera staleness, and cross-view
consistency~\cite{li2026liberovpro}.
LIBERO-Pro2.0 further covers static and runtime shifts across
environmental, observational, execution, robot-state, and language
dimensions~\cite{zhou2026liberopro2}.
RoboPRO complements success-based evaluation with collision
metrics under clutter and controlled
perturbations~\cite{li2026robopro}.
Building on these evaluation principles, our work focuses on
mirror-augmented manipulation, where reflections introduce
structured visual ambiguity alongside physical scene changes.
We evaluate task outcomes and contact behavior, and use
geometrically matched non-reflective panels to examine
performance differences beyond the presence of the panel itself.

\subsection{Mirror Perception and Spatial Reasoning}

Mirrors challenge visual understanding because reflected objects
can resemble directly observed objects while occupying different
physical locations.
Early mirror-detection methods identify reflective regions through
contextual contrasts~\cite{yang2019mirror}, while video-based
approaches exploit spatial and temporal
correspondences~\cite{lin2023videomirror}.
These methods address the localization of mirror regions rather
than the action consequences of reflected observations.
Mirror3D further estimates mirror geometry to refine unreliable depth measurements on reflective surfaces~\cite{Tan_2021_CVPR}.
More recently, MIRA-Bench evaluates reflection-aware perception,
spatial reasoning, and scene-level decisions in real RGB-D mirror
scenes, exposing difficulties in reconciling reflected appearance
with depth measurements~\cite{yin2026mira}.
MirrorQA examines subject-centered left--right reasoning in
mirror contexts~\cite{liu2026mirrorqa}.
Related diagnostic benchmarks, including
HallusionBench~\cite{guan2024hallusionbench} and
IllusionVQA~\cite{shahgir2024illusionvqa}, investigate visual
illusions and failures of image-grounded reasoning.
Unlike these perception-focused studies, we evaluate task outcomes and contact behavior during closed-loop manipulation under reflections.

\subsection{Mirror-Mediated Robotic Interaction}

Robotic mirror research spans geometry and sensorimotor learning.
Hart and Scassellati infer mirror viewpoints from learned sensorimotor relationships~\cite{hart2012mirror}.
Marchand and Chaumette develop visual servoing for robot-mounted mirrors~\cite{marchand2017mirror}, while Lanillos et al. combine active inference and sensorimotor prediction for mirror self-recognition and self/other distinction~\cite{lanillos2020mirror}.
Yoshioka et al. exploit mirror reflections to observe
occluded regions, while UMI incorporates side mirrors
for manipulation and shows that corrected mirror views
can outperform uncorrected ones~\cite{yoshioka2021lookingglassdiminishingocclusions,chi2024universalmanipulationinterfaceinthewild}.
Recent embodied evaluations extend mirror-based tests to
multimodal agents.
MirrorBench studies self-centric intelligence through simulated
mirror interaction~\cite{guo2026mirrorbench}, while
Ziliotto et al.\ evaluate whether agents can infer hidden
self-attributes from reflections and use this evidence to guide
actions~\cite{ziliotto2026mirror}.
Beyond self-centric tasks, Being-M0.7 evaluates mirror-assisted
grasping of a directly occluded toy in a real-world humanoid
task~\cite{yue2026beingm07}.
Our work studies reflection robustness across multiple
generalist policies and manipulation tasks under controlled
mirror configurations.
\section{Benchmark Design and Evaluation Protocol}
\label{sec:benchmark}

\begin{figure*}[t]
    \centering
    \includegraphics[width=\textwidth]{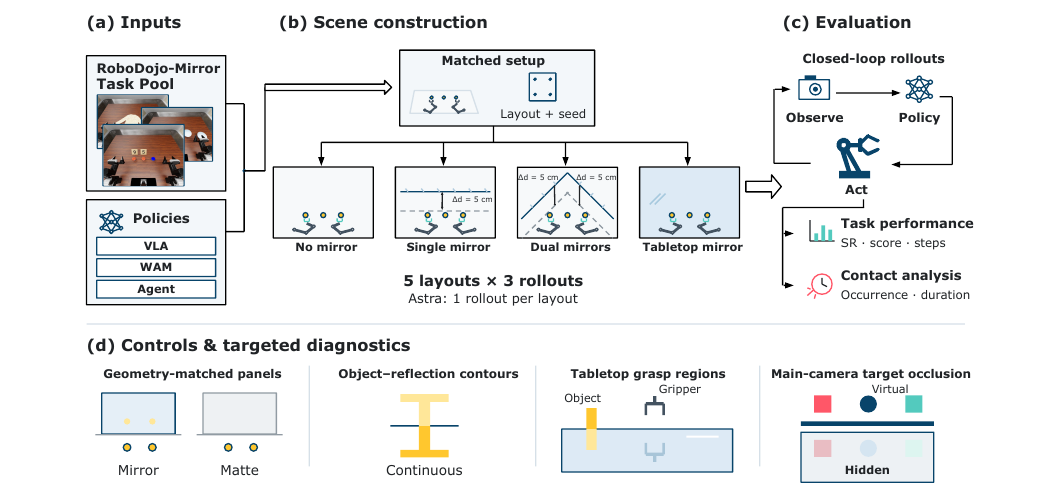}
    \caption{Overview of the evaluation protocol in RoboDojo-Mirror. Policies are evaluated on matched task layouts under four scene conditions, with additional non-reflective panel controls for selected policies.}
    \label{fig:benchmark}
\end{figure*}

Built on RoboDojo~\cite{chen2026robodojo}, RoboDojo-Mirror evaluates policy robustness to reflections while preserving the original task objectives and success criteria. Fig.~\ref{fig:benchmark} summarizes the paired evaluation protocol and controls.

\subsection{Policy and Task Selection}
We evaluate five policies spanning Vision-Language-Action models, World Action Models, and agentic policies: WALL-OSS-1.0-Lite (WALL-OSS)~\cite{xsquare2026wallosslite}, VPP2~\cite{guo2026videopredictionpolicy2}, Awomo-0.5~\cite{team2026awomo}, Rex-M1 Preview~\cite{futian2026rexm1preview}, and GPT-6-Astra (Astra)~\cite{zhang2026unexpectedrobotpolicyearly, openai2026gpt6astra}. This selection covers different approaches to generating robot actions from visual observations.

For open-weight policies, we use official checkpoints and inference configurations. 
Within each policy, we hold the checkpoint, observation interface, action interface, and inference settings fixed across scene conditions. 
For Astra, the available camera streams, robot-state interface, motion tools, and inference budget are likewise held fixed across conditions.

We select a shared eight-task set $\mathcal{T}$ that balances diverse manipulation requirements with strong overall RoboDojo leaderboard performance.
All policies are evaluated on $\mathcal{T}$ under four conditions, using paired comparisons against their own mirror-free baselines. The tasks are listed in the supplementary material.

\subsection{Mirror Configurations}
\label{sec:sec3_2}
We construct four scene conditions from each task layout: no mirror, a single vertical mirror, dual vertical mirrors, and a mirrored tabletop. These conditions vary the placement and geometry of reflections while retaining matched initial object and robot states.

\textbf{No Mirror.} We retain the original RoboDojo scene as the baseline.

\textbf{Single Mirror.} A vertical planar mirror is placed near a tabletop edge, with its plane parallel to that edge and its reflective side facing the manipulation area.

\textbf{Dual Mirrors.} Two vertical mirrors are placed along adjacent sides of the workspace, with mutually perpendicular planes and reflective faces oriented toward the manipulation area. This configuration introduces reflections from both mirrors and may also show higher-order reflections, depending on visibility and rendering settings.

\textbf{Mirrored Tabletop.} We replace the tabletop’s surface material with a reflective finish while retaining its position and dimensions. Reflections therefore appear near the objects and the surface on which manipulation takes place.

For the single-mirror and dual-mirror conditions, we shift the panels by 5 cm away from the robot relative to their initial construction positions, providing additional clearance while retaining visible reflections. The panels are fixed in the scene. This placement does not guarantee collision-free execution; contact with the panels is therefore recorded separately from task success.

For WALL-OSS and Awomo-0.5, we also replace the
single and dual mirrors with geometry-matched
non-reflective panels as physical-obstruction
controls. No corresponding control is used for
the mirrored tabletop.

Across the four main evaluation conditions, we retain the original task instructions, matched initial object and robot states, and task success criteria. During scene validation, we inspect the camera streams supplied to each policy to check reflection visibility, direct target visibility, and initial intersections between the inserted panels and existing scene geometry. We retain the scene and camera configurations with the evaluation records. The target-occlusion experiments in Section~\ref{sec:sec4_3} use a separate diagnostic setup.

\subsection{Evaluation Protocol and Metrics}
\textbf{Paired Evaluation.} For each policy and selected task, we run separate closed-loop episodes with matched layouts and policy seeds across scene conditions. The main evaluation uses five layouts and three policy seeds per task for WALL-OSS, VPP2, Awomo-0.5, and Rex-M1 Preview. Astra uses five layouts with one rollout per layout.

\textbf{Termination and budgets.} We retain the original task success criteria and task-specific termination rules. An episode ends when the task terminates or its execution budget is exhausted. Panel contact is recorded but is not an additional termination condition. For Astra, the execution budget also includes a limit on model calls. Diagnostic experiments report their own layouts, seeds, and budgets when these differ from the main evaluation.

Let $y_{m,t,c,l,s}\in\{0,1\}$ denote task success
for policy $m$, task $t$, condition $c$, layout $l$,
and policy-seed index $s$.
With evaluated layout and seed sets
$\mathcal L_{m,t}$ and $\mathcal S_m$, task SR
and overall SR are
\begin{equation}
    \begin{aligned}
        \mathrm{SR}_{m,t,c}
        &=\frac{100}{|\mathcal L_{m,t}||\mathcal S_m|}
        \sum_{l\in\mathcal L_{m,t}}\sum_{s\in\mathcal S_m}
        y_{m,t,c,l,s},\\
        \mathrm{SR}_{m,c}
        &=\frac{1}{|\mathcal T|}
        \sum_{t\in\mathcal T}\mathrm{SR}_{m,t,c}.
    \end{aligned}
\end{equation}
Thus, SR is expressed as a percentage, with equal
weight for each task in the fixed eight-task set
$\mathcal T$.
Relative to the no-mirror baseline $B$,
\begin{equation}
    \Delta\mathrm{SR}_{m,c}
    =\mathrm{SR}_{m,c}-\mathrm{SR}_{m,B}.
\end{equation}
Negative changes indicate degradation.
The corresponding drop is $-\Delta\mathrm{SR}_{m,c}$.
Table~\ref{tab:main_results} averages drops equally
over the three mirror conditions $\mathcal M$:
\begin{equation}
    D_m=-\frac{1}{3}\sum_{c\in\mathcal M}
    \Delta\mathrm{SR}_{m,c}.
\end{equation}
Changes and drops are in percentage points;
baseline comparisons use matched layouts and
policy-seed indices within each task and policy.
Native task scores, scaled to 0--100, are aggregated
in the same way to capture partial completion.

We report 95\% layout-cluster bootstrap intervals,
resampling layouts within fixed tasks while retaining
all seeds and condition pairing
(details in supplementary material).

\textbf{Contact and behavior logging.} We record visual observations, robot states, and action trajectories throughout each rollout. For episodes with valid contact measurements, we additionally monitor physical interactions with the mirror panel. We define the contact episode rate as the proportion of episodes involving at least one recorded contact with the panel.
\section{Performance under Mirror Conditions}
\label{sec:performance}

Following Section~\ref{sec:benchmark}, we report overall performance, mirror--panel comparisons, targeted behavioral diagnostics, and real-world validation.
Trajectory, video, and contact analyses characterize three observed behaviors: collision, phantom targeting, and spatial mislocalization.

\textbf{Collision.} The robot contacts the mirror during execution, potentially disrupting manipulation. We distinguish such contact events from task failures, as contact does not necessarily cause failure.

\textbf{Phantom targeting.} The robot directs manipulation actions toward a reflected object rather than its physical counterpart, suggesting that the reflection is treated as an actionable target.

\textbf{Spatial mislocalization.} The robot acts at an incorrect location relative to the physical target, resulting in missed grasps or ineffective manipulation without clear evidence of phantom targeting.

\subsection{Overall Performance}
\label{sec:sec4_1}
Table~\ref{tab:main_results} reports task scores and success rates aggregated over the task set, with each policy compared against its own mirror-free baseline.

All five policies achieve mirror-free success rates of 60.8--72.5\%, and every mirror configuration reduces their aggregate success rates.
The average paired SR drops range from 10.8 to 56.7 percentage points across policies.
With a mirrored tabletop, VPP2 drops from 72.5\% to 27.5\%, while Rex-M1 Preview drops from 66.7\% to 4.2\%. Task scores also decrease under all mirror configurations.

Overall, dual mirrors tend to cause greater performance degradation than single mirrors, with VPP2 as an exception. WALL-OSS and Awomo-0.5 reach their lowest success rates under dual mirrors, whereas VPP2, Rex-M1 Preview, and Astra perform worst with a mirrored tabletop.

Astra shows little aggregate change under a single mirror (65.0\% to 62.5\%), but its SR decreases to 52.5\% under dual mirrors and 47.5\% with a mirrored tabletop. Despite the highest recorded robot--mirror contact rates under single and dual mirrors (55.0\% and 85.0\%), Astra exhibits relatively small SR drops, suggesting that contact occurrence alone does not explain performance degradation.

Since execution budgets differ across policies, these aggregates should not be interpreted as a controlled cross-policy ranking. 
Furthermore, task outcomes alone cannot separate visual interference from physical obstruction, motivating the mirror--panel comparisons in the next subsection.

\begin{table*}[t]
\centering
\caption{Task performance under mirror perturbations. Scene-condition entries report Score / SR (\%). Parenthesized percentages in the single- and dual-mirror columns indicate the fraction of episodes with at least one recorded robot--mirror contact. Drops average baseline-minus-mirror differences equally across the three mirror conditions, using matched layouts and policy-seed indices. Brackets indicate 95\% paired layout-cluster bootstrap confidence intervals for SR drops.}
\label{tab:main_results}
\small
\setlength{\tabcolsep}{3pt}
\renewcommand{\arraystretch}{1.15}
\begin{tabular*}{\textwidth}{@{\extracolsep{\fill}}lcccccc@{}}
\toprule
& \multicolumn{4}{c}{Scene condition: Score / SR (\%)} & \multicolumn{2}{c}{Average paired drop} \\
\cmidrule(lr){2-5}\cmidrule(l){6-7}
Policy & None & Single & Dual & Tabletop & Score $\downarrow$ & SR (pp) $\downarrow$ \\
\midrule
WALL-OSS & 70.6 / 67.5 & 23.3 / 21.7 (34.2\%) & 17.2 / 15.0 (28.3\%) & 27.2 / 25.0 & 48.0 & \shortstack{46.9\\{\scriptsize [38.3, 55.3]}} \\
VPP2 & 73.2 / 72.5 & 36.8 / 33.3 (49.2\%) & 40.9 / 35.0 (47.5\%) & 34.3 / 27.5 & 35.8 & \shortstack{40.6\\{\scriptsize [34.2, 46.9]}} \\
Awomo-0.5 & 62.7 / 60.8 & 17.8 / 15.8 (27.5\%) & 8.0 / 6.7 (30.0\%) & 13.0 / 10.0 & 49.8 & \shortstack{50.0\\{\scriptsize [42.2, 58.1]}} \\
Rex-M1 Preview & 68.8 / 66.7 & 18.0 / 15.8 (12.5\%) & 13.2 / 10.0 (45.0\%) & 6.7 / 4.2 & 56.2 & \shortstack{56.7\\{\scriptsize [48.3, 65.0]}} \\
Astra & 66.3 / 65.0 & 63.3 / 62.5 (55.0\%) & 57.0 / 52.5 (85.0\%) & 51.3 / 47.5 & 9.1 & \shortstack{10.8\\{\scriptsize [1.7, 20.0]}} \\
\bottomrule
\end{tabular*}
\end{table*}

\subsection{Mirror--Panel Comparisons}
\label{sec:sec4_2}
Mirrors introduce both visual reflections and physical obstacles, with frequent robot--mirror contact observed in Table \ref{tab:main_results}. To examine whether direct robot--surface contact alone can explain performance degradation, we compare mirrors with geometry-matched non-reflective panels (Section \ref{sec:sec3_2}).

\textbf{Non-contact Failures and Spatial Mislocalization.} Let \(h\in\{0,1\}\) indicate whether robot–panel contact occurs. Omitting other indices for brevity, we define two sets of paired episodes:
\begin{equation}
    \begin{aligned}
        \mathcal{F} &: y_{\mathrm{NoMirror}}=y_{\mathrm{Panel}}=1,\quad y_{\mathrm{Mirror}}=0 \\
        \mathcal{N} &: y_{\mathrm{NoMirror}}=1,\quad h_{\mathrm{Mirror}}=h_{\mathrm{Panel}}=0
    \end{aligned}
\end{equation}
Here, \(\mathcal F\) contains mirror-specific failures, while \(\mathcal N\) contains pairs in which the no-mirror baseline succeeds and neither the mirror nor the panel condition involves robot–panel contact.

Table \ref{tab:mirror_panel} shows that mirror-only failures outnumber panel-only failures across all 120 matched pairs per setting, particularly under dual mirrors (25 / 2 for WALL-OSS and 42 / 3 for Awomo-0.5). This asymmetry persists within the non-contact subset \(\mathcal{N}\).
Moreover, 50.0--87.5\% of mirror-specific failures in \(\mathcal{F}\) occur without recorded robot--panel contact in either condition. These observations indicate failure differences beyond direct collisions, consistent with possible spatial mislocalization of physical targets.

\begin{table}[h]
    \centering
    \caption{Matched mirror--panel comparisons under single-
    and dual-mirror configurations. ALL reports mirror-only / panel-only failure counts across 120 matched pairs per setting. For $\mathcal{F}$, values
    report the number and percentage of mirror-specific failures
    without robot--panel contact in either condition.
    For $\mathcal{N}$, values report mirror-only versus
    panel-only failure counts.}
    \label{tab:mirror_panel}
    \footnotesize
    \setlength{\tabcolsep}{5pt}
    \renewcommand{\arraystretch}{1.0}
    \begin{tabular}{@{}llcc@{}}
        \toprule
        Subset & Config. & WALL-OSS & Awomo-0.5 \\
        \midrule
        ALL & Single & 9 / 4 & 16 / 7 \\
            & Dual   & 25 / 2 & 42 / 3 \\
        \midrule
        $\mathcal{F}$
            & Single & 7 / 8 (87.5\%) & 9 / 13 (69.2\%) \\
            & Dual   & 11 / 22 (50.0\%) & 20 / 34 (58.8\%) \\
        \midrule
        $\mathcal{N}$
            & Single & 7 / 2 & 9 / 5 \\
            & Dual   & 11 / 1 & 20 / 2 \\
        \bottomrule
    \end{tabular}
\end{table}

\textbf{Persistent Contact Behavior and Phantom Targeting. }Beyond non-contact failures, we investigate whether reflections alter robot behavior during physical contact. We define the set of paired episodes in which both conditions involve robot–panel contact:
\begin{equation}
\mathcal{C} : y_{\mathrm{NoMirror}}=1,\quad h_{\mathrm{Mirror}}=h_{\mathrm{Panel}}=1
\end{equation}
Within \(\mathcal{C}\) under dual mirrors, panel success rates exceed mirror success rates by 29.4 and 41.2 percentage points for WALL-OSS and Awomo-0.5, indicating different outcomes despite contact in both conditions. 

To characterize contact persistence, we define the Contact Time Fraction as
\begin{equation}
R_{\mathrm{contact}}=\frac{T_{\mathrm{contact}}}{T_{\mathrm{execution}}}
\end{equation}
Within \(\mathcal F\cap\mathcal C\) under dual mirrors, mean contact time fractions are 1.85 and 6.83 times higher with mirrors than panels for WALL-OSS ($n=6$) and Awomo-0.5 ($n=7$), respectively. 
This behavior is consistent with possible phantom targeting, where a policy may repeatedly attempt to reach the apparent location of a reflected object, only to be obstructed by the physical mirror. 
In contrast, the single-mirror configuration does not exhibit similarly pronounced and consistent differences in contact behavior. 
Complete statistics are provided in the supplementary material.

Overall, the mirror--panel comparisons reveal failures beyond direct collisions,
with behavioral patterns consistent with spatial mislocalization and phantom targeting. Section \ref{sec:sec4_3} further investigates these failure modes through controlled scenarios and execution videos.

\subsection{Targeted Behavioral Diagnostics}
\label{sec:sec4_3}

To complement aggregate task outcomes, we use three
diagnostics to characterize grasping behavior,
pushing progress, and physical-target interaction
under different relationships between objects
and their reflections.

\subsubsection{Grasping on a Mirrored Tabletop}

A mirrored tabletop places reflected object and robot
features near directly visible targets.
We examine whether policies grasp the physical object
or close and lift without securing it.

\begin{figure}[h]
    \centering
    \includegraphics[width=\linewidth]{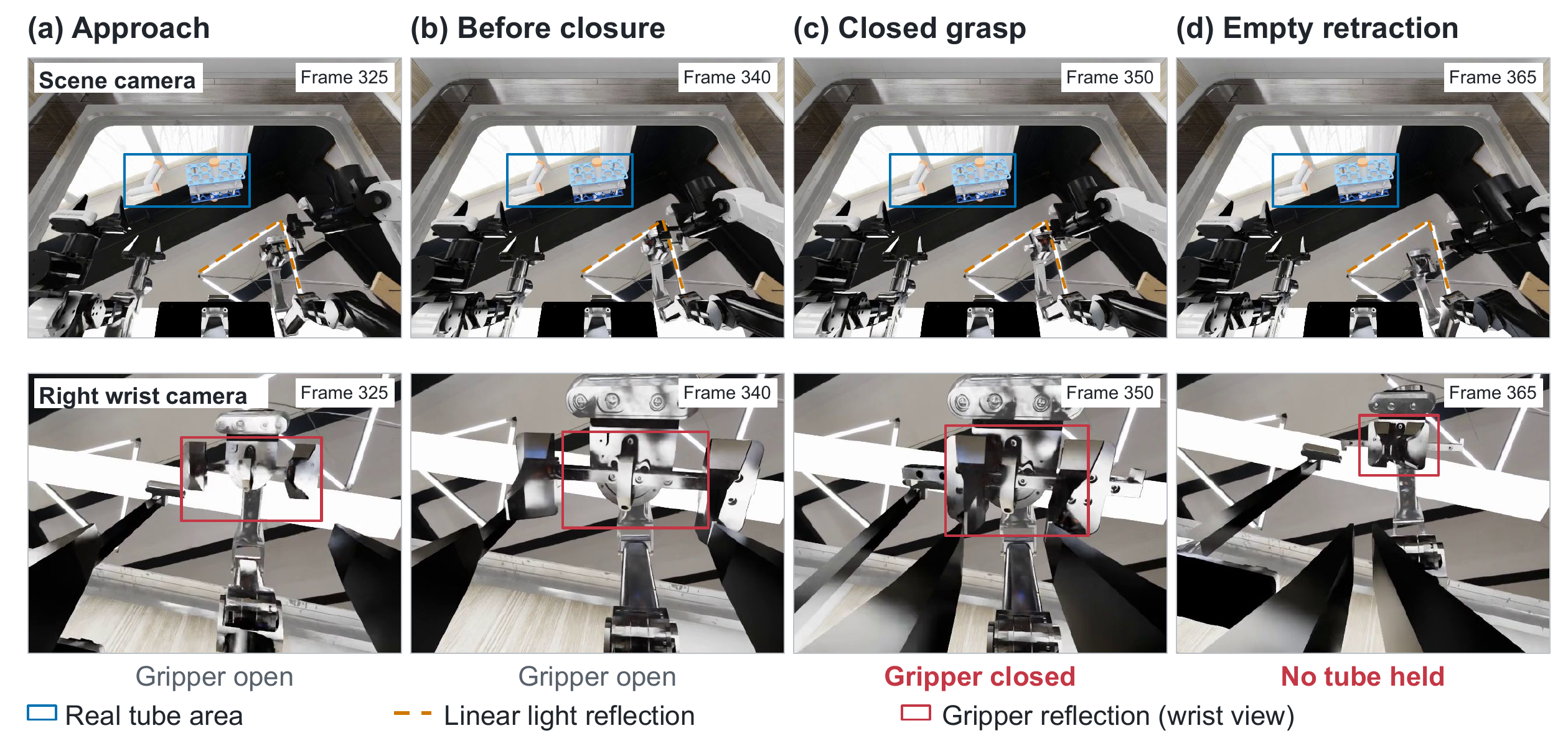}
    \caption{Empty-grasp behavior consistent with ECL-G
    on a mirrored tabletop. Scene and right wrist-camera
    views show closing and lifting without securing
    the physical tube. Annotations distinguish the
    tube from nearby reflections.}
    \label{fig:empty_grasp}
\end{figure}

Fig.~\ref{fig:empty_grasp} illustrates behavior
consistent with geometric empty close-and-lift
(ECL-G), quantified below: the gripper closes
away from the physical tube and lifts empty-handed.
The closing-location mismatch is consistent with
spatial mislocalization in this example.

Table~\ref{tab:tabletop_ecl_0} compares VPP2,
Awomo-0.5, WALL-OSS, and Astra on tube insertion,
block stacking, and bottle disposal.
Conditions are paired by layout within each policy,
with 45 episodes per condition (15 per task),
or 15 for Astra (five per task).
Beyond task success, we examine whether closing
and lifting the gripper actually moves the object.
We report three episode-level rates:
geometric empty close-and-lift (ECL-G), where the
gripper closes and lifts with no target in its
sensing region; no lift (NL), where a closing
interval lacks the required gripper lift; and
no confirmed object following (NF), where a lift
occurs but the object is not confirmed to move
with the gripper.
These events may overlap within an episode.

Mirrored tabletops increase ECL-G and NF for all
four policies.
VPP2's SR falls from 97.8\% to 4.4\%, while
ECL-G rises from 0.0\% to 95.6\%:
almost every mirrored episode contains closing
and lifting without enclosing the physical target.
For Astra, SR remains unchanged and NL decreases,
yet ECL-G and NF increase.
Thus, fewer no-lift events do not necessarily
mean better grasping, and unchanged task success
can hide more frequent empty or unconfirmed lifts.

\begin{table}[t]
    \centering
    \caption{Task success and grasp-related behaviors
    pooled across three tasks. Entries are
    No mirror/Mirrored tabletop percentages.
    Each condition contains 45 episodes per policy,
    except Astra (15). Behavioral events may overlap.}
    \label{tab:tabletop_ecl_0}
    \footnotesize
    \setlength{\tabcolsep}{4pt}
    \renewcommand{\arraystretch}{1.1}
    \begin{tabular*}{\linewidth}
        {@{\extracolsep{\fill}}lcccc@{}}
        \toprule
        Policy & Task SR & ECL-G & NL & NF \\
        \midrule
        VPP2
        & 97.8/4.4 & 0.0/95.6
        & 26.7/42.2 & 6.7/100.0 \\
        Awomo-0.5
        & 80.0/26.7 & 8.9/17.8
        & 35.6/62.2 & 17.8/44.4 \\
        WALL-OSS
        & 60.0/28.9 & 24.4/40.0
        & 42.2/53.3 & 26.7/51.1 \\
        Astra
        & 33.3/33.3 & 13.3/26.7
        & 53.3/46.7 & 60.0/73.3 \\
        \bottomrule
    \end{tabular*}
\end{table}

\subsubsection{Pushing an Object Connected to Its Reflection}

\begin{figure}[h]
    \centering
    \includegraphics[width=\linewidth]{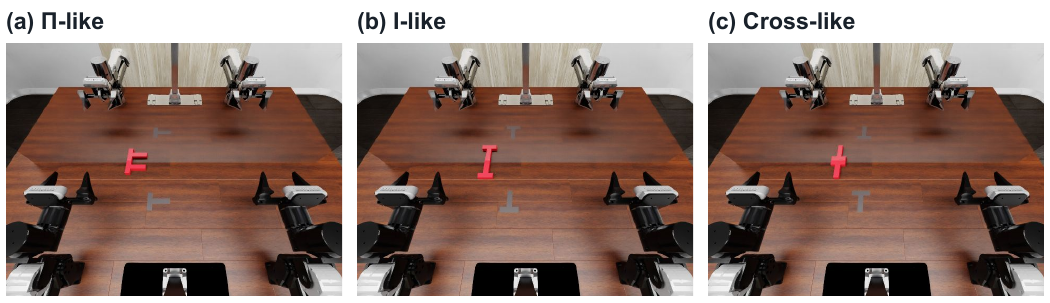}
    \caption{Composite contours formed by a physical T-shaped
    object adjoining its mirror reflection:
    (a) sideways $\Pi$-like, (b) I-like, and (c) cross-like.}
    \label{fig:push_t_single_mirror_cvpr_v3}
\end{figure}

In the preceding grasping example, the physical target and nearby reflections remain separately visible. 
We next examine pushing when a physical object adjoins its reflection.
In a mirror-based Push-T variant~\cite{florence2021implicit,chi2023diffusionpolicy}, the object and its reflection form $\Pi$-like, I-like, or cross-like contours (Fig.~\ref{fig:push_t_single_mirror_cvpr_v3}).

We evaluate VPP2, WALL-OSS, and Astra on five matched
placements per contour.
VPP2 and WALL-OSS use three seeds per placement
($n=15$), while Astra uses one rollout per placement
($n=5$).
Episodes end upon sustained valid separation
or after 1,000 control steps.
We examine whether policies can adjust their actions
to move the physical T-shaped object away from its
reflection and break the connected contour.
Table~\ref{tab:push_t_behavior} reports the rate
of reaching 10\,mm displacement, maximum displacement,
steps to this threshold, and robot--mirror contact counts.
Higher reaching rates and larger displacements indicate
more frequent and greater object movement.
Fewer steps indicate earlier threshold crossing
among episodes that reach it.
More mirror contacts combined with little object
movement suggest difficulty adjusting actions
to achieve effective pushing.

\begin{table}[t]
    \centering
    \caption{Pushing progress and robot--mirror contacts.
    Displacement and steps: mean $\pm$ sample SD;
    steps include only episodes reaching 10\,mm
    (``--'': none).
    Contacts: mean events per episode.}
    \label{tab:push_t_behavior}
    \footnotesize
    \setlength{\tabcolsep}{3pt}
    \renewcommand{\arraystretch}{1.0}
    \begin{tabular}{@{}lcccc@{}}
        \toprule
        \textbf{Contour}
        & \shortstack{\textbf{Reached}\\\textbf{10 mm (\%)}}
        & \shortstack{\textbf{Max. disp.}\\\textbf{(mm)}}
        & \shortstack{\textbf{Steps to}\\\textbf{10 mm}}
        & \shortstack{\textbf{Mirror}\\\textbf{contacts}} \\
        \midrule
        \multicolumn{5}{@{}l}{
            \textit{VPP2 (WAM), $n=15$}} \\
        Cross-like
        & 53.3 & $23.64 \pm 23.38$
        & $308.1 \pm 216.7$ & 13.07 \\
        $\Pi$-like
        & 0 & $<0.01$ & -- & 11.80 \\
        I-like
        & 93.3 & $57.79 \pm 48.53$
        & $303.0 \pm 168.7$ & 6.27 \\
        \midrule
        \multicolumn{5}{@{}l}{
            \textit{WALL-OSS (VLA), $n=15$}} \\
        Cross-like
        & 86.7 & $47.11 \pm 31.25$
        & $258.3 \pm 70.0$ & 13.87 \\
        $\Pi$-like
        & 80.0 & $32.84 \pm 44.07$
        & $420.9 \pm 192.5$ & 17.20 \\
        I-like
        & 100 & $105.97 \pm 41.90$
        & $105.6 \pm 22.7$ & 6.47 \\
        \midrule
        \multicolumn{5}{@{}l}{
            \textit{Astra (agentic), $n=5$}} \\
        Cross-like
        & 100 & $135.04 \pm 42.04$
        & $229.6 \pm 111.9$ & 5.60 \\
        $\Pi$-like
        & 100 & $174.64 \pm 5.32$
        & $208.2 \pm 156.5$ & 8.40 \\
        I-like
        & 100 & $160.69 \pm 15.06$
        & $232.4 \pm 224.8$ & 4.20 \\
        \bottomrule
    \end{tabular}
\end{table}

Astra achieves sustained valid separation in all
15 episodes, with mean maximum displacements of
135.04--174.64\,mm and 4.20--8.40 mirror contacts
per episode, compared with 6.27--17.20 for the
other two policies.
Its repeated trajectory changes after mirror
contact suggest effective adjustment to the mirror
constraint.
VPP2 and WALL-OSS make the greatest progress under
I-like contours, reaching 10\,mm in 93.3\% and
100\% of episodes, respectively.
Under $\Pi$-like contours, VPP2 moves less than
0.01\,mm despite averaging 11.80 mirror contacts.
These results highlight policy-dependent adjustment
when separating a physical object from its
adjoining reflection.

\subsubsection{Acting with Reflections as the Main Visual Cue}
\label{sec:target_occlusion}

\begin{figure}[h]
    \centering
    \includegraphics[width=\linewidth]{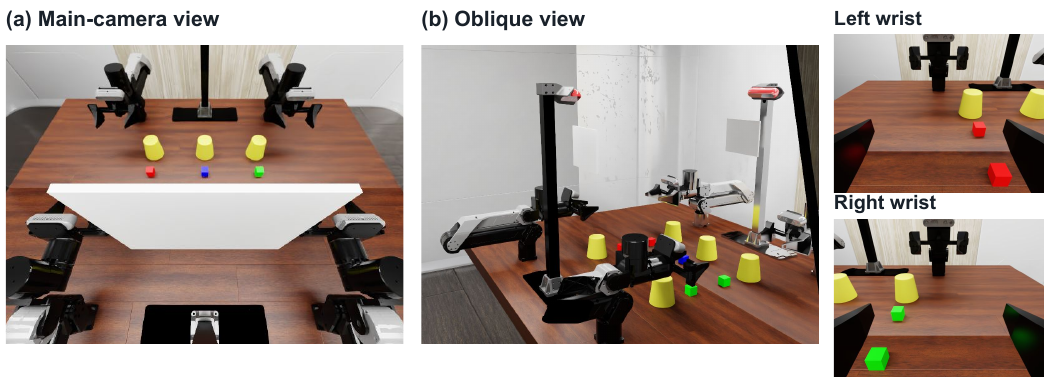}
    \caption{
    Target occlusion in \texttt{cover\_blocks}.
    The panel hides physical targets from the main camera
    while preserving their reflections; some targets
    remain directly visible to wrist cameras.
    }
    \label{fig:floating_panel_setup_v2}
\end{figure}

We test whether policies can locate and manipulate
physical targets when the main camera sees their
reflections but not the targets directly, and whether
explicit mirror-related guidance helps.
In \texttt{cover\_blocks}, an elevated panel blocks
the main camera's direct view of the blocks and cups
while preserving their mirror reflections
(Fig.~\ref{fig:floating_panel_setup_v2}).
Wrist cameras retain direct visibility of some targets.
We measure physical-target contact and task success
to assess whether policies reach the real objects
and complete the task under this restricted view.

We compare WALL-OSS, Awomo-0.5, and Astra under
Original (unchanged instruction), Brief (mirror hint),
and Detailed (scene description) prompts.
Each prompt uses the same five layouts, with three
seeds per layout for WALL-OSS and Awomo-0.5
($n=15$), and one rollout per layout for Astra ($n=5$).
Table~\ref{tab:reflection_prompt_contact} reports
physical-target contact rates and native task SR.
Contact rates measure how often the robot touches
real cups or blocks; higher rates indicate more
frequent physical interaction.
Cup contact is directly relevant because the task
requires moving cups to cover blocks.
SR measures whether these interactions result in
task completion.
Comparing both metrics across prompts tests whether
mirror-related guidance helps policies reach and
successfully manipulate the physical targets.

\begin{table}[t]
    \centering
    \caption{Target contact and task success under
    main-camera occlusion. All values are percentages.}
    \label{tab:reflection_prompt_contact}
    \footnotesize
    \setlength{\tabcolsep}{3pt}
    \renewcommand{\arraystretch}{1.0}
    \begin{tabular*}{\linewidth}
        {@{\extracolsep{\fill}}llrrr@{}}
        \toprule
        Policy & Prompt & Cup contact & Block contact & SR \\
        \midrule
        WALL-OSS
        & Original & 20.0 & 20.0 & 0.0 \\
        & Brief    & 20.0 & 40.0 & 0.0 \\
        & Detailed &  0.0 & 13.3 & 0.0 \\
        \midrule
        Awomo-0.5
        & Original & 20.0 & 0.0 & 0.0 \\
        & Brief    & 20.0 & 0.0 & 0.0 \\
        & Detailed & 26.7 & 0.0 & 0.0 \\
        \midrule
        Astra
        & Original & 100.0 & 0.0 & 20.0 \\
        & Brief    & 100.0 & 0.0 &  0.0 \\
        & Detailed & 100.0 & 0.0 & 20.0 \\
        \bottomrule
    \end{tabular*}
\end{table}

WALL-OSS and Awomo-0.5 contact physical cups in only
0.0--26.7\% of episodes and never complete the task,
indicating difficulty reaching and manipulating
the task-relevant objects under main-camera occlusion.
For WALL-OSS, Brief raises block contact from
20.0\% to 40.0\% without improving cup contact or SR.
Astra reaches cups in every episode, but completes
the task in only 1/5 episodes under Original and
Detailed, and 0/5 under Brief.
Thus, reaching physical targets remains distinct
from successfully manipulating them, and mirror-related
guidance provides no consistent improvement in
these small samples.

\subsection{Real-World Validation}

To examine whether the performance degradation observed in simulation extends to physical manipulation, we evaluate robot policies under real-world mirror conditions. 

\begin{figure}[tbp]
    \centering
    \includegraphics[width=\linewidth]{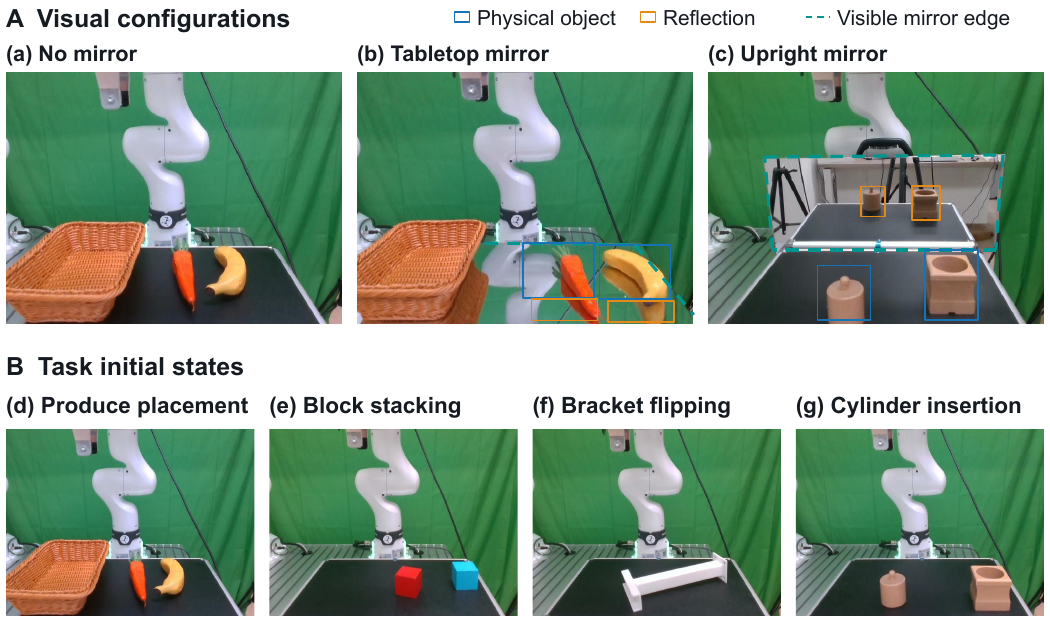}
    \caption{Real-world visual conditions and task initial states. 
    (A) No-mirror, mirrored-tabletop, and upright-mirror configurations; annotations identify physical objects, reflections, and the visible mirror edge. 
    (B) Initial states for produce placement, block stacking, bracket flipping, and cylinder insertion.}
    \label{fig:real_robot_mirror_cvpr}
\end{figure}

We evaluate $\pi_{0.5}$~\cite{intelligence2025pi05}
and FastWAM~\cite{yuan2026fastwam} on a Franka robot
across four tasks: produce placement, block stacking,
bracket flipping, and cylinder insertion.
Each task uses no-mirror, upright-mirror,
and mirrored-tabletop conditions
(Fig.~\ref{fig:real_robot_mirror_cvpr}).
The upright condition uses a single vertical mirror,
matching the simulated single-mirror configuration in type.
Five matched initial layouts with two repetitions each
yield 10 trials per task, policy, and condition.
Checkpoints, instructions, camera configurations,
and execution budgets remain fixed across conditions.
Further experimental details are provided
in the supplementary material.

Table~\ref{tab:real_world_results} shows lower
overall SR under both mirror conditions.
For $\pi_{0.5}$, SR decreases from 77.5\% without
mirrors to 52.5\% with an upright mirror and
27.5\% with a mirrored tabletop;
FastWAM decreases from 60.0\% to 30.0\% and
10.0\%, respectively.
Both policies achieve 80.0\% on block stacking
without mirrors but 0.0\% with a mirrored tabletop.
These results show an overall degradation trend
consistent with simulation.

\begin{table}[t]
    \centering
    \caption{
    Real-world success rates (\%) under different
    mirror conditions. Each setting includes
    10 trials (five layouts and two repetitions per layout).
    }
    \label{tab:real_world_results}
    \footnotesize
    \setlength{\tabcolsep}{2pt}
    \renewcommand{\arraystretch}{1.0}
    \begin{tabular*}{\linewidth}
        {@{\extracolsep{\fill}}lrrrrrr@{}}
        \toprule
        & \multicolumn{3}{c}{$\pi_{0.5}$}
        & \multicolumn{3}{c}{FastWAM} \\
        \cmidrule(lr){2-4}
        \cmidrule(l){5-7}
        \textbf{Task}
        & None & Upright & Tabletop
        & None & Upright & Tabletop \\
        \midrule
        Produce placement
        & 90.0 & 90.0 & 60.0
        & 80.0 & 60.0 & 40.0 \\
        Block stack
        & 80.0 & 30.0 & 0.0
        & 80.0 & 0.0 & 0.0 \\
        Bracket flip
        & 90.0 & 70.0 & 50.0
        & 60.0 & 40.0 & 0.0 \\
        Cylinder insert
        & 50.0 & 20.0 & 0.0
        & 20.0 & 20.0 & 0.0 \\
        \midrule
        \textbf{Overall}
        & \textbf{77.5} & \textbf{52.5} & \textbf{27.5}
        & \textbf{60.0} & \textbf{30.0} & \textbf{10.0} \\
        \bottomrule
    \end{tabular*}
\end{table}

\section{Conclusion}
\label{sec:conclusion}

We introduce RoboDojo-Mirror to evaluate robotic
manipulation under controlled mirror configurations.
Paired evaluations across five policies and real-world
experiments show lower success rates under mirrors.
Matched panel controls show that direct collisions alone cannot explain the performance degradation. Behavioral diagnostics reveal patterns consistent with phantom targeting and spatial mislocalization.
These findings expose difficulties in connecting
visual observations to effective physical actions
and establish reflection robustness as a concrete
target for evaluating and improving robot policies.

\section*{Acknowledgments}
This research was supported in part with Cloud TPUs from
Google's TPU Research Cloud (TRC).

{
    \small
    \bibliographystyle{ieeenat_fullname}
    \bibliography{main}

@String(CVPR= {IEEE Conf. Comput. Vis. Pattern Recog.})

@String(AAAI = {AAAI})

@String(CVPR  = {CVPR})

@inproceedings{liu2023libero,
  title     = {{LIBERO}: Benchmarking Knowledge Transfer for Lifelong Robot Learning},
  author    = {Liu, Bo and Zhu, Yifeng and Gao, Chongkai and Feng, Yihao and Liu, Qiang and Zhu, Yuke and Stone, Peter},
  booktitle = {Advances in Neural Information Processing Systems},
  volume    = {36},
  pages     = {44776--44791},
  year      = {2023},
  url       = {https://proceedings.neurips.cc/paper_files/paper/2023/hash/8c3c666820ea055a77726d66fc7d447f-Abstract-Datasets_and_Benchmarks.html}
}

@inproceedings{chen2026robotwin2,
  title     = {{RoboTwin 2.0}: A Scalable Data Generator and Benchmark with Strong Domain Randomization for Robust Bimanual Robotic Manipulation},
  author    = {Chen, Tianxing and Chen, Zanxin and Chen, Baijun and Cai, Zijian and Liu, Yibin and Li, Zixuan and Liang, Qiwei and Lin, Xianliang and Ge, Yiheng and Gu, Zhenyu and Deng, Weiliang and Guo, Yubin and Nian, Tian and Xie, Xuanbing and Chen, Qiangyu and Su, Kailun and Xu, Tianling and Liu, Guodong and Hu, Mengkang and Gao, Huan-Ang and Wang, Kaixuan and Liang, Zhixuan and Qin, Yusen and Yang, Xiaokang and Luo, Ping and Mu, Yao},
  booktitle = {Proceedings of the International Conference on Machine Learning},
  volume    = {306},
  series    = {Proceedings of Machine Learning Research},
  pages     = {13673--13699},
  year      = {2026},
  url       = {https://proceedings.mlr.press/v306/chen26j.html}
}

@article{chen2026robodojo,
  title   = {{RoboDojo}: A Unified Sim-and-Real Benchmark for Comprehensive Evaluation of Generalist Robot Manipulation Policies},
  author  = {Chen, Tianxing and Chen, Yue and Li, Zixuan and Tang, Junyuan and Su, Kailun and Lu, Haoran and Wan, Weijie and Chen, Baijun and Liu, Songling and Yan, Haowen and Su, Honghao and Dou, Zhiyang and Wang, Kaixuan and Zhang, Dandan and Liu, Yunze and Qin, Yan and Liang, Qiwei and Wu, Qiwei and Lin, Zijian and Lin, Wenwei and Wang, Yuran and He, Minghua and Wu, Tianshu and Wu, Ruihai and Zhou, Jingquan and Lei, Kai-Chong and Yu, Haibao and Ji, Yuanfeng and Jin, Weiyang and Lin, Guanyu and Li, Xiaofan and Xiong, Qi and Xu, Renjing and Li, Zhongyu and Chai, Wenhao and Xie, Enze and Wang, Ziwei and Mu, Yao and Dong, Hao and Matusik, Wojciech and Ding, Mingyu and Ding, Wenbo and Luo, Ping and Tomizuka, Masayoshi},
  journal = {arXiv preprint arXiv:2607.04434},
  year    = {2026},
  url     = {https://arxiv.org/abs/2607.04434}
}

@inproceedings{fei2026liberoplus,
  title     = {{LIBERO-Plus}: A Progressive Robustness Benchmark for Visual-Language-Action Models},
  author    = {Fei, Senyu and Wang, Siyin and Shi, Junhao and Dai, Zihao and Cai, Jikun and Qian, Pengfang and Ji, Li and He, Xinzhe and Zhang, Shiduo and Fei, Zhaoye and Fu, Jinlan and Gong, Jingjing and Qiu, Xipeng},
  booktitle = {Proceedings of the IEEE/CVF Conference on Computer Vision and Pattern Recognition},
  pages     = {38574--38583},
  year      = {2026},
  url       = {https://openaccess.thecvf.com/content/CVPR2026/html/Fei_LIBERO-Plus_A_Progressive_Robustness_Benchmark_for_Visual-Language-Action_Models_CVPR_2026_paper.html}
}

@article{li2026liberovpro,
  title   = {{LIBERO-VPro}: Benchmarking Closed-Loop Visual Robustness of Robotic Foundation Models},
  author  = {Li, Huiqiong and Mei, Zhiting and Majumdar, Anirudha and Chen, Jingjing and Jiang, Yu-Gang and Zhu, Bin},
  journal = {arXiv preprint arXiv:2609.24350},
  year    = {2026},
  url     = {https://arxiv.org/abs/2609.24350}
}

@article{zhou2026liberopro2,
  title   = {{LIBERO-Pro2.0}: A Benchmark for Evaluating the Robustness of Embodied Foundation Models},
  author  = {Zhou, Xueyang and Xu, Yangming and Li, Qianjiang and Tie, Guiyao and Wu, Junhao and Huang, Ziheng and Zhang, Guowen and Yuan, Zhijie and Wu, Di and Chu, Duanfeng and Zhou, Pan and Sun, Lichao and Chen, Yongchao},
  journal = {arXiv preprint arXiv:2610.24587},
  year    = {2026},
  url     = {https://arxiv.org/abs/2610.24587}
}

@inproceedings{li2026robopro,
  title     = {Data and Evaluation Protocols for Robust Robotic Manipulation},
  author    = {Li, Zhiyuan and Eret, Charles and Zhao, Xuan and Wu, Yangzheng and Rasouli, Amir},
  booktitle = {Data-Centric Robotics Workshop at Robotics: Science and Systems},
  year      = {2026},
  url       = {https://robopro-bench.github.io/RoboPRO/}
}

@inproceedings{yang2019mirror,
  title     = {Where Is My Mirror?},
  author    = {Yang, Xin and Mei, Haiyang and Xu, Ke and Wei, Xiaopeng and Yin, Baocai and Lau, Rynson W. H.},
  booktitle = {Proceedings of the IEEE/CVF International Conference on Computer Vision},
  pages     = {8809--8818},
  year      = {2019},
  url       = {https://openaccess.thecvf.com/content_ICCV_2019/html/Yang_Where_Is_My_Mirror_ICCV_2019_paper.html}
}

@inproceedings{lin2023videomirror,
  title     = {Learning To Detect Mirrors From Videos via Dual Correspondences},
  author    = {Lin, Jiaying and Tan, Xin and Lau, Rynson W. H.},
  booktitle = {Proceedings of the IEEE/CVF Conference on Computer Vision and Pattern Recognition},
  pages     = {9109--9118},
  year      = {2023},
  url       = {https://openaccess.thecvf.com/content/CVPR2023/html/Lin_Learning_To_Detect_Mirrors_From_Videos_via_Dual_Correspondences_CVPR_2023_paper.html}
}

@article{yin2026mira,
  title   = {When Depth Lies: Benchmarking Vision-Language Models on Mirror-Induced {RGB-D} Ambiguity},
  author  = {Yin, Hao and Guo, Tianchen and Du, Heming and Ke, Yan and Liu, Yanbin and Yu, Xin},
  journal = {arXiv preprint arXiv:2610.41651},
  year    = {2026},
  url     = {https://arxiv.org/abs/2610.41651}
}

@inproceedings{liu2026mirrorqa,
  title     = {{MirrorQA}: Benchmarking Multimodal {LLMs} on Mirror-Orientation Reasoning},
  author    = {Liu, Jingping and Peng, Xingchen and Zhou, Yan and Liu, Ziyan and Zhai, Jie and Chen, Ronghao and Wang, Huacan and Jia, Xiaofeng},
  booktitle = {Proceedings of the 64th Annual Meeting of the Association for Computational Linguistics (Volume 1: Long Papers)},
  pages     = {40464--40476},
  year      = {2026},
  doi       = {10.18653/v1/2026.acl-long.1879},
  url       = {https://aclanthology.org/2026.acl-long.1879/}
}

@inproceedings{guan2024hallusionbench,
  title     = {{HallusionBench}: An Advanced Diagnostic Suite for Entangled Language Hallucination and Visual Illusion in Large Vision-Language Models},
  author    = {Guan, Tianrui and Liu, Fuxiao and Wu, Xiyang and Xian, Ruiqi and Li, Zongxia and Liu, Xiaoyu and Wang, Xijun and Chen, Lichang and Huang, Furong and Yacoob, Yaser and Manocha, Dinesh and Zhou, Tianyi},
  booktitle = {Proceedings of the IEEE/CVF Conference on Computer Vision and Pattern Recognition},
  pages     = {14375--14385},
  year      = {2024},
  url       = {https://openaccess.thecvf.com/content/CVPR2024/html/Guan_HallusionBench_An_Advanced_Diagnostic_Suite_for_Entangled_Language_Hallucination_and_CVPR_2024_paper.html}
}

@inproceedings{shahgir2024illusionvqa,
  title     = {{IllusionVQA}: A Challenging Optical Illusion Dataset for Vision Language Models},
  author    = {Shahgir, Haz Sameen and Sayeed, Khondker Salman and Bhattacharjee, Abhik and Ahmad, Wasi Uddin and Dong, Yue and Shahriyar, Rifat},
  booktitle = {Proceedings of the Conference on Language Modeling},
  year      = {2024},
  url       = {https://openreview.net/forum?id=7ysaJGs7zY}
}

@inproceedings{hart2012mirror,
  title     = {Mirror Perspective-Taking with a Humanoid Robot},
  author    = {Hart, Justin W. and Scassellati, Brian},
  booktitle = {Proceedings of the AAAI Conference on Artificial Intelligence},
  volume    = {26},
  number    = {1},
  pages     = {1990--1996},
  year      = {2012},
  doi       = {10.1609/aaai.v26i1.8389},
  url       = {https://ojs.aaai.org/index.php/AAAI/article/view/8389}
}

@inproceedings{marchand2017mirror,
  title     = {Visual Servoing through Mirror Reflection},
  author    = {Marchand, Eric and Chaumette, Francois},
  booktitle = {IEEE International Conference on Robotics and Automation},
  year      = {2017},
  doi       = {10.1109/ICRA.2017.7989439},
  url       = {https://hal.inria.fr/hal-01445484}
}

@inproceedings{lanillos2020mirror,
  title     = {Robot Self/Other Distinction: Active Inference Meets Neural Networks Learning in a Mirror},
  author    = {Lanillos, Pablo and Pages, Jordi and Cheng, Gordon},
  booktitle = {Proceedings of the European Conference on Artificial Intelligence},
  pages     = {2410--2416},
  year      = {2020},
  doi       = {10.3233/FAIA200372},
  url       = {https://arxiv.org/abs/2004.05473}
}

@article{guo2026mirrorbench,
  title   = {{MirrorBench}: Evaluating Self-centric Intelligence in {MLLMs} by Introducing a Mirror},
  author  = {Guo, Shengyu and Ye, Tongrui and Zhang, Jianbo and Zhang, Zicheng and Li, Chunyi and Zhai, Guangtao},
  journal = {arXiv preprint arXiv:2604.14785},
  year    = {2026},
  url     = {https://arxiv.org/abs/2604.14785}
}

@article{ziliotto2026mirror,
  title   = {Mirror, Mirror on the Wall: Can {VLM} Agents Tell Who They Are at All?},
  author  = {Ziliotto, Filippo and Beneduce, Ciro and Lepri, Bruno and Serafini, Luciano and Luca, Massimiliano and Campari, Tommaso},
  journal = {arXiv preprint arXiv:2605.08816},
  year    = {2026},
  url     = {https://arxiv.org/abs/2605.08816}
}

@article{gallup1970selfrecognition,
  author = {Gallup, Jr., Gordon G.},
  title = {Chimpanzees: Self-Recognition},
  journal = {Science},
  volume = {167},
  number = {3914},
  pages = {86--87},
  year = {1970},
  doi = {10.1126/science.167.3914.86},
  url = {https://doi.org/10.1126/science.167.3914.86}
}

@inproceedings{BlackK-RSS-25,
  author = {Kevin Black and Noah Brown and Danny Driess and Adnan Esmail and Michael Robert Equi and Chelsea Finn and Niccolo Fusai and Lachy Groom and Karol Hausman and Brian Ichter and Szymon Jakubczak and Tim Jones and Liyiming Ke and Sergey Levine and Adrian Li-Bell and Mohith Mothukuri and Suraj Nair and Karl Pertsch and Lucy Xiaoyang Shi and Laura Smith and James Tanner and Quan Vuong and Anna Walling and Haohuan Wang and Ury Zhilinsky},
  title = {{$\pi_0$}: A Vision-Language-Action Flow Model for General Robot Control},
  booktitle = {Proceedings of Robotics: Science and Systems},
  year = {2025},
  address = {Los Angeles, CA, USA},
  month = jun,
  doi = {10.15607/RSS.2025.XXI.010},
  url = {https://www.roboticsproceedings.org/rss21/p010.html}
}

@misc{ye2026worldactionmodelszeroshot,
  title = {World Action Models are Zero-shot Policies},
  author = {Seonghyeon Ye and Yunhao Ge and Kaiyuan Zheng and Shenyuan Gao and Sihyun Yu and George Kurian and Suneel Indupuru and You Liang Tan and Chuning Zhu and Jiannan Xiang and Ayaan Malik and Kyungmin Lee and William Liang and Nadun Ranawaka and Jiasheng Gu and Yinzhen Xu and Guanzhi Wang and Fengyuan Hu and Avnish Narayan and Johan Bjorck and Jing Wang and Gwanghyun Kim and Dantong Niu and Ruijie Zheng and Yuqi Xie and Jimmy Wu and Qi Wang and Ryan Julian and Danfei Xu and Yilun Du and Yevgen Chebotar and Scott Reed and Jan Kautz and Yuke Zhu and Linxi {Jim} Fan and Joel Jang},
  year = {2026},
  eprint = {2602.15922},
  archivePrefix = {arXiv},
  primaryClass = {cs.RO},
  url = {https://arxiv.org/abs/2602.15922}
}

@misc{jia2026agentpolicyroboticmanipulation,
  title = {Agent as Policy for Robotic Manipulation},
  author = {Mengzhao Jia and Yang Lin and Xixin Zhang and Zhihan Zhang and Xiaobai Liu and Meng Jiang},
  year = {2026},
  eprint = {2609.12541},
  archivePrefix = {arXiv},
  primaryClass = {cs.CL},
  url = {https://arxiv.org/abs/2609.12541}
}

@misc{xsquare2026wallosslite,
  author = {{X Square Robot}},
  title = {{WALL-OSS 1.0 Lite} --- {RoboDojo}},
  year = {2026},
  howpublished = {Hugging Face model card. \url{https://huggingface.co/x-square-robot/wall-oss-1.0-lite-robodojo}},
  url = {https://huggingface.co/x-square-robot/wall-oss-1.0-lite-robodojo},
  note = {Accessed November 13, 2026}
}

@misc{guo2026videopredictionpolicy2,
  title = {Video Prediction Policy 2: Predict Better, Act Better},
  author = {Yanjiang Guo and Haodong Yan and Zhide Zhong and Zhongru Zhang and Qingyuan Yang and Qingzhou Lu and Xiaoyu Chen and Yen-Jen Wang and Shuying Deng and Chenghan Yang and Puzhen Yuan and Chenxin Liu and Tun Ban and Xiang Zhu and Yichen Liu and Kun Feng and Haoang Li and Jianyu Chen},
  year = {2026},
  eprint = {2610.10270},
  note = {arXiv:2610.10270},
  archivePrefix = {arXiv},
  primaryClass = {cs.CV},
  url = {https://arxiv.org/abs/2610.10270}
}

@techreport{team2026awomo,
  title = {{Awomo-0.5} Technical Report},
  author = {{Awomo-WAM Team}},
  institution = {Awomo},
  year = {2026},
  note = {\url{https://github.com/Awomo-WestlakeDI/Awomo-0.5/blob/main/Awomo_0_5.pdf}},
  url = {https://github.com/Awomo-WestlakeDI/Awomo-0.5/blob/main/Awomo_0_5.pdf}
}

@misc{futian2026rexm1preview,
  title = {{Rex-M1 Preview}: {XPolicyLab/RoboDojo} Policy Release},
  author = {{Futian Lab} and {VisIncept}},
  year = {2026},
  howpublished = {XPolicyLab release documentation. \url{https://github.com/XPolicyLab/XPolicyLab/blob/main/policy/Rex_M1_preview/README.md}},
  url = {https://github.com/XPolicyLab/XPolicyLab/blob/main/policy/Rex_M1_preview/README.md},
  note = {Accessed November 13, 2026}
}

@misc{openai2026gpt6astra,
  title = {{GPT-6 Astra}: A New Generation of Intelligence},
  author = {{OpenAI}},
  year = {2026},
  howpublished = {Official model announcement. \url{https://openai.com/index/gpt-6-astra/}},
  url = {https://openai.com/index/gpt-6-astra/},
  note = {Accessed November 13, 2026}
}

@misc{zhang2026unexpectedrobotpolicyearly,
      title={An Unexpected Robot Policy: Early Evaluations of GPT-6 Astra on RoboDojo and Beyond}, 
      author={Wenbo Zhang and Kaixuan Wang and Yutao Ouyang and Xiaoyu Huang and Liyang Li and Kailun Su and Weiyang Jin and Wenhao Chai and Haotian Liang and Zhiyang Dou and Yue Chen and Tianxing Chen},
      year={2026},
      eprint={2609.24170},
      archivePrefix={arXiv},
      primaryClass={cs.CV},
      url={https://arxiv.org/abs/2609.24170}, 
}

@misc{florence2021implicit,
  title = {Implicit Behavioral Cloning},
  author = {Pete Florence and Corey Lynch and Andy Zeng and Oscar Ramirez and Ayzaan Wahid and Laura Downs and Adrian Wong and Johnny Lee and Igor Mordatch and Jonathan Tompson},
  year = {2021},
  eprint = {2109.00137},
  note = {arXiv:2109.00137},
  archivePrefix = {arXiv},
  primaryClass = {cs.RO},
  url = {https://arxiv.org/abs/2109.00137}
}

@inproceedings{chi2023diffusionpolicy,
  title = {Diffusion Policy: Visuomotor Policy Learning via Action Diffusion},
  author = {Chi, Cheng and Feng, Siyuan and Du, Yilun and Xu, Zhenjia and Cousineau, Eric and Burchfiel, Benjamin and Song, Shuran},
  booktitle = {Proceedings of Robotics: Science and Systems},
  year = {2023},
  url = {https://diffusion-policy.cs.columbia.edu/}
}

@misc{intelligence2025pi05,
  title = {{$\pi_{0.5}$}: A Vision-Language-Action Model with Open-World Generalization},
  author = {{Physical Intelligence} and Kevin Black and Noah Brown and James Darpinian and Karan Dhabalia and Danny Driess and Adnan Esmail and Michael Equi and Chelsea Finn and Niccolo Fusai and Manuel Y. Galliker and Dibya Ghosh and Lachy Groom and Karol Hausman and Brian Ichter and Szymon Jakubczak and Tim Jones and Liyiming Ke and Devin LeBlanc and Sergey Levine and Adrian Li-Bell and Mohith Mothukuri and Suraj Nair and Karl Pertsch and Allen Z. Ren and Lucy Xiaoyang Shi and Laura Smith and Jost Tobias Springenberg and Kyle Stachowicz and James Tanner and Quan Vuong and Homer Walke and Anna Walling and Haohuan Wang and Lili Yu and Ury Zhilinsky},
  year = {2025},
  eprint = {2504.16054},
  note = {arXiv:2504.16054},
  archivePrefix = {arXiv},
  primaryClass = {cs.LG},
  url = {https://arxiv.org/abs/2504.16054}
}

@misc{yuan2026fastwam,
  title = {{Fast-WAM}: Do World Action Models Need Test-time Future Imagination?},
  author = {Tianyuan Yuan and Zibin Dong and Yicheng Liu and Hang Zhao},
  year = {2026},
  eprint = {2603.16666},
  note = {arXiv:2603.16666},
  archivePrefix = {arXiv},
  primaryClass = {cs.CV},
  url = {https://arxiv.org/abs/2603.16666}
}

@article{yue2026beingm07,
  title   = {{Being-M0.7}: A Latent World-Action Model for Humanoid Robots},
  author  = {Yue, Junpeng and Li, Boyuan and Wang, Yuxuan and Wang, Zepeng and Fu, Yuhui and Xie, Feiyang and Zhang, Yu and Zhang, Jing and Zhang, Xianqi and Li, Weibo and Zheng, Xiaofei and Fang, Yuming and Wang, Jiangxing and Lu, Zongqing},
  journal = {arXiv preprint arXiv:2610.11283},
  year    = {2026},
  url     = {https://arxiv.org/abs/2610.11283}
}

@InProceedings{Tan_2021_CVPR,
    author    = {Tan, Jiaqi and Lin, Weijie and Chang, Angel X. and Savva, Manolis},
    title     = {Mirror3D: Depth Refinement for Mirror Surfaces},
    booktitle = {Proceedings of the IEEE/CVF Conference on Computer Vision and Pattern Recognition (CVPR)},
    month     = {June},
    year      = {2021},
    pages     = {15990-15999}
}

@misc{yoshioka2021lookingglassdiminishingocclusions,
      title={Through the Looking Glass: Diminishing Occlusions in Robot Vision Systems with Mirror Reflections}, 
      author={Kentaro Yoshioka and Hidenori Okuni and Tuan Thanh Ta and Akihide Sai},
      year={2021},
      eprint={2108.13599},
      archivePrefix={arXiv},
      primaryClass={cs.RO},
      url={https://arxiv.org/abs/2108.13599}, 
}

@misc{chi2024universalmanipulationinterfaceinthewild,
      title={Universal Manipulation Interface: In-The-Wild Robot Teaching Without In-The-Wild Robots}, 
      author={Cheng Chi and Zhenjia Xu and Chuer Pan and Eric Cousineau and Benjamin Burchfiel and Siyuan Feng and Russ Tedrake and Shuran Song},
      year={2024},
      eprint={2402.10329},
      archivePrefix={arXiv},
      primaryClass={cs.RO},
      url={https://arxiv.org/abs/2402.10329}, 
}
}

\clearpage
\renewcommand{\dblfloatpagefraction}{0.75}
\setcounter{page}{1}
\maketitlesupplementary

\appendix

\section*{Overview}
\label{sec:rationale}

This supplementary material documents scene geometry,
policy configurations, and evaluation metrics
(Appendix~\ref{supp:evaluation_details}); task-level
results (Appendix~\ref{supp:task_results}); mirror--panel
controls and contact measurements
(Appendix~\ref{supp:panel_details}); tabletop grasping,
Push-T, and target-occlusion diagnostics
(Appendices~\ref{supp:tabletop}, \ref{supp:push_t}, and
\ref{supp:occlusion}); and real-world deployment details
(Appendix~\ref{supp:real_world}).

\section{Scene Geometry and Evaluation Details}
\label{supp:evaluation_details}

\subsection{Mirror Geometry and Virtual Images}
\label{supp:mirror_geometry}

\begin{figure}[b]
    \centering
    \includegraphics[width=\linewidth]
        {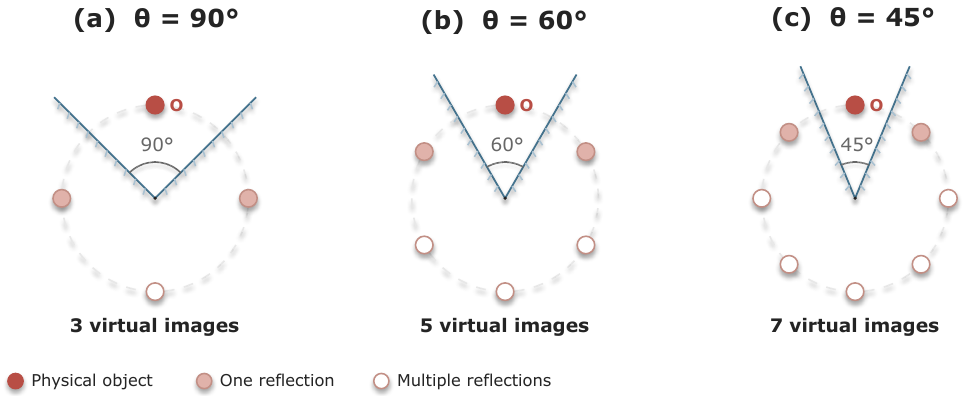}
    \caption{
        Idealized virtual-image arrangements for two
        mirrors with included angles of
        (a) $90^\circ$, (b) $60^\circ$, and
        (c) $45^\circ$.
        Solid red markers denote the physical object;
        filled pale markers denote first-order images;
        outlined markers denote higher-order images.
        The diagrams illustrate reflection geometry,
        not camera-visible image counts.
    }
    \label{fig:mirror_angle_images}
\end{figure}

Two mirrors can produce both first-order images,
formed by a single reflection, and higher-order
images, formed by successive reflections between
the mirrors.
Figure~\ref{fig:mirror_angle_images} illustrates
the virtual-image arrangements for three idealized
mirror configurations.
The illustrated $90^\circ$, $60^\circ$, and
$45^\circ$ configurations produce three, five,
and seven virtual images, respectively.

Our main dual-mirror evaluation uses perpendicular
mirrors ($90^\circ$).
The other angles illustrate reflection geometry
rather than additional evaluated conditions.
In the experimental scenes, finite mirror dimensions,
camera viewpoints, occlusion, and rendering settings
determine which virtual images are actually visible
in the policy observations.

\subsection{Policy Configurations}
\label{supp:policy_configurations}

We evaluate four pretrained robot policies:
WALL-OSS-1.0-Lite, VPP2, Awomo-0.5, and
Rex-M1 Preview, together with the tool-using
GPT-6-Astra agent.
For the four pretrained policies, we use their
officially released checkpoints and corresponding
inference configurations, without additional
training or mirror-specific adaptation.

Within each policy, we keep the checkpoint,
observation preprocessing, camera inputs,
action representation, and inference settings
fixed across scene conditions.
Task instructions and native success criteria
are likewise preserved.
The scene condition is not supplied as an
additional policy input in the main evaluation.
The mirror-related instruction variants are
restricted to the target-occlusion diagnostic.

Policy-specific inference and control settings
are retained rather than forced into a common
action interface.
Consequently, comparisons primarily measure
changes relative to each policy's own
mirror-free baseline.
GPT-6-Astra additionally uses an explicit
robot-tool interface and a decision-call budget,
as detailed below.

Table~\ref{supp:tab:policy_configurations} summarizes
the policy configurations.
Prediction horizon denotes the number of action
steps generated per inference, while replanning
interval denotes the number executed before
requesting a new prediction.

\begin{table*}[tp]
    \centering
    \caption{
        Policy configurations used in evaluation.
        H+W denotes the head and both wrist cameras.
        EE denotes end-effector pose and gripper targets;
        Joint denotes joint-position and gripper targets.
        Horizons and replanning intervals are measured
        in control steps.
        Precision describes policy inference, not
        simulator arithmetic.
    }
    \label{supp:tab:policy_configurations}
    \footnotesize
    \setlength{\tabcolsep}{4pt}
    \renewcommand{\arraystretch}{1.2}
    \begin{tabular*}{\textwidth}{
        @{\extracolsep{\fill}}llccccc@{}
    }
        \toprule
        Policy & Checkpoint / model
        & RGB cameras & Action
        & \shortstack{Prediction\\horizon}
        & \shortstack{Replanning\\interval}
        & Precision \\
        \midrule
        WALL-OSS-1.0-Lite
        & \shortstack[l]{\texttt{wall-oss-1.0-lite-}\\
                         \texttt{robodojo}}
        & H+W & EE & 32 & 16
        & BF16/FP32 \\

        VPP2
        & \texttt{joint2b\_s100000}
        & H+W & EE & 32 & 24
        & BF16 \\

        Awomo-0.5
        & \shortstack[l]{RoboDojo release\\
                         \texttt{model.pt}}
        & H+W & Joint & 16 & 8
        & BF16 \\

        Rex-M1 Preview
        & \shortstack[l]{\texttt{Rex\_M1\_preview}\\
                         \texttt{last.ckpt}}
        & H+W & EE & 30 & 15
        & BF16 \\

        GPT-6-Astra
        & \texttt{gpt-6-astra}
        & H+W & EE tools & Variable & Tool-dependent
        & Not disclosed \\
        \bottomrule
    \end{tabular*}
\end{table*}

WALL-OSS uses selective BF16 parameter casting
while retaining FP32 components.
Awomo-0.5 uses the original RoboDojo checkpoint.
VPP2 uses the RoboDojo EE16 deployment adapter;
the checkpoint name does not indicate the
executed action representation.
GPT-6-Astra selects individual motion-tool actions
with variable execution duration, rather than
predicting fixed-length action chunks.
Its reasoning effort is set to \texttt{medium},
which is distinct from numerical precision.

\subsubsection{GPT-6-Astra Agent Configuration}
\label{supp:astra_configuration}

We configure GPT-6-Astra as a closed-loop,
tool-using agent with the RoboDojo Inspect-EEF
interface.
We access the model through the Codex CLI's
\texttt{app-server} interface using ChatGPT
subscription authentication, rather than
direct, API-key-authenticated access to the
OpenAI API.
The model is requested as \texttt{gpt-6-astra}
with reasoning effort set to \texttt{medium}.
At each observation boundary, it receives RGB
images from the head camera and both wrist cameras,
robot proprioceptive state, and execution feedback.
Depth observations and simulator ground-truth
object coordinates are not provided.

The agent uses two tools.
\texttt{move\_eef} specifies world-frame
end-effector position, orientation, and gripper
opening for one or both arms.
The native executor validates the targets,
plans joint trajectories with cuRobo, and executes
them through the simulator's robot controller.
The agent subsequently receives execution feedback
and updated observations.
The \texttt{give\_up} tool allows the agent to
terminate an unsuccessful attempt.

We retain the upstream Inspect-EEF prompt and
task-specific recipes.
These recipes provide static task descriptions
and scoring criteria; they do not expose live
object states, current scores, or success flags.
No additional demonstrations, previous-attempt
trajectories, or human guidance are supplied
during an episode.

Each episode starts a fresh Codex session.
Subsequent decisions continue the same session
with new observations and execution feedback.
Within-episode history is managed by the native
session, with no cross-episode history transfer.
Although the local policy buffer retains two
observation groups, this does not impose a
two-observation limit on the model-visible
session history.

Each episode permits at most 100 agent decision
calls and remains subject to the native task's
control-step budget and termination rules.
The decision-call limit is distinct from the
number of low-level control steps.
We keep the agent configuration, tool interface,
and budgets fixed across scene conditions.
Task outcomes are determined by the native
environment evaluator.

\subsection{Experiment Coverage and Measurement Sources}

Table~\ref{supp:tab:experiment_coverage} summarizes the
sample structure and measurement sources of the
reported experiments.
A cell fixes the policy, task (or Push-T contour),
and experimental condition; layouts and repetitions
are counted within that cell.
Scene conditions, instruction variants, and
collision/patch settings define separate cells.
Layouts are paired within a policy and task;
the same layout count does not imply identical
initial states across policies.

\begin{table*}[tp]
    \centering
    \caption{Sample structure and measurement sources.
    $n$ counts original policy executions per cell.}
    \label{supp:tab:experiment_coverage}
    \footnotesize
    \setlength{\tabcolsep}{4pt}
    \renewcommand{\arraystretch}{1.12}
    \begin{tabular}{@{}p{0.22\textwidth}p{0.23\textwidth}cccp{0.29\textwidth}@{}}
        \toprule
        Experiment and scope & Policies
        & \shortstack{Layouts\\per cell}
        & \shortstack{Runs per\\layout} & $n$
        & Measurement source \tabularnewline
        \midrule
        \raggedright Main evaluation: eight tasks, four scenes & \raggedright WALL-OSS, VPP2, Awomo-0.5, Rex-M1 Preview & 5 & 3 & 15 & \raggedright Native task evaluator; policy-specific contact records \tabularnewline
         & \raggedright Astra & 5 & 1 & 5 & \raggedright Native task evaluator and recorded executions \tabularnewline
        \midrule
        \raggedright Mirror--panel controls: eight tasks, two panel settings & \raggedright WALL-OSS, Awomo-0.5 & 5 & 3 & 15 & \raggedright Native task evaluator; physics-substep robot--surface contacts \tabularnewline
        \midrule
        \raggedright Tabletop diagnostic: three tasks, two conditions & \raggedright VPP2, Awomo-0.5, WALL-OSS & 5 & 3 & 15 & \raggedright Native outcomes; recorded trajectories and supplementary simulator measurements \tabularnewline
         & \raggedright Astra & 5 & 1 & 5 & \raggedright Native outcomes; original trajectories for 20 episodes and simulator measurements for 10 \tabularnewline
        \midrule
        \raggedright Push-T patch/collision controls: three contours, four settings & \raggedright VPP2 & 1 & 5 & 5 & \raggedright Object trajectories, projected silhouettes, and robot--mirror contact records \tabularnewline
        \midrule
        \raggedright Push-T placement diagnostic: three contours & \raggedright VPP2, WALL-OSS & 5 & 3 & 15 & \raggedright Object trajectories, projected silhouettes, and robot--mirror contact records \tabularnewline
         & \raggedright Astra & 5 & 1 & 5 & \raggedright Object trajectories, projected silhouettes, and robot--mirror contact records \tabularnewline
        \midrule
        \raggedright Target occlusion: one task, three instructions & \raggedright WALL-OSS, Awomo-0.5 & 5 & 3 & 15 & \raggedright Native outcomes; simulator-measured target contacts from recorded actions \tabularnewline
         & \raggedright Astra & 5 & 1 & 5 & \raggedright Native outcomes; simulator-measured target contacts from recorded actions \tabularnewline
        \midrule
        \raggedright Real-world evaluation: four tasks, three scenes & \raggedright $\pi_{0.5}$, FastWAM & 5 & 2 & 10 & \raggedright Recorded task outcomes; policy-related execution failures included \tabularnewline
        \bottomrule
    \end{tabular}
\end{table*}

\subsection{Paired Evaluation and Metrics}
\label{supp:paired_metrics}

Each main-evaluation task--condition cell contains
five layouts and three policy seeds for WALL-OSS,
VPP2, Awomo-0.5, and Rex-M1 Preview (15 episodes),
or five layouts with one execution per layout for
GPT-6-Astra (five episodes).
Matching is within policy and task, rather than
across policies.

The native evaluator assigns each episode a binary success
indicator and a task score in $[0,1]$ that captures partial
completion. For each policy, task, and scene condition,
we report the success rate as a percentage and scale the
mean task score to $0$--$100$.
For each policy and scene condition, we macro-average both
metrics across the eight tasks, giving each task equal weight.
Mirror-free baseline performance can vary across policies
on the same task. The task-level tables report these
policy-specific baselines for all eight tasks.

For a given metric $v$ (score or success rate), let
$\bar v_{m,t,c}$ denote its mean for policy $m$ on task $t$
under scene condition $c$. The main table reports the average
paired drop from the no-mirror baseline $B$ across the eight
tasks in $\mathcal T$ and the three mirror conditions in
$\mathcal M$:
\begin{equation}
D_m(v)=\frac{1}{3|\mathcal T|}
\sum_{c\in\mathcal M}\sum_{t\in\mathcal T}
\left(\bar v_{m,t,B}-\bar v_{m,t,c}\right).
\end{equation}
Positive values indicate lower performance under mirror
conditions. Success-rate drops are reported in percentage
points.

We report 95\% percentile confidence intervals for the
average success-rate drop using 20{,}000 paired bootstrap
replicates. In each replicate, we sample five layouts with
replacement within each task. We retain all runs associated
with each sampled layout and use the same layout draws
across scene conditions to preserve pairing. The intervals
thus capture variation across layouts for the evaluated
set of tasks.

An execution reaching its task budget remains an
evaluated episode; contact is not an extra termination
condition. We count only original policy executions as trials.
For the Astra batches, preserved technical interruptions
are not independently scored as task failures; accepted
completed recoveries occupy their original evaluation
slots and are not counted as extra repetitions.

\section{Task Coverage and Detailed Results}
\label{supp:task_results}

\subsection{Common Task Set}

We selected the eight RoboDojo tasks in
Table~\ref{supp:tab:task_list} to cover diverse manipulation
requirements while favoring tasks with strong overall
performance on the RoboDojo leaderboard.
The tasks span object pickup and rearrangement, stacking,
insertion, cloth folding, and button pressing in response
to a numerical instruction.
In our mirror-free evaluations, at least three of the five
policies achieved a success rate of 60\% or higher on seven
of the eight tasks. Baseline performance varied across
policies for individual tasks.

\begin{figure*}[p]
    \centering
    \includegraphics[
        width=\textwidth,
        height=0.84\textheight,
        keepaspectratio
    ]{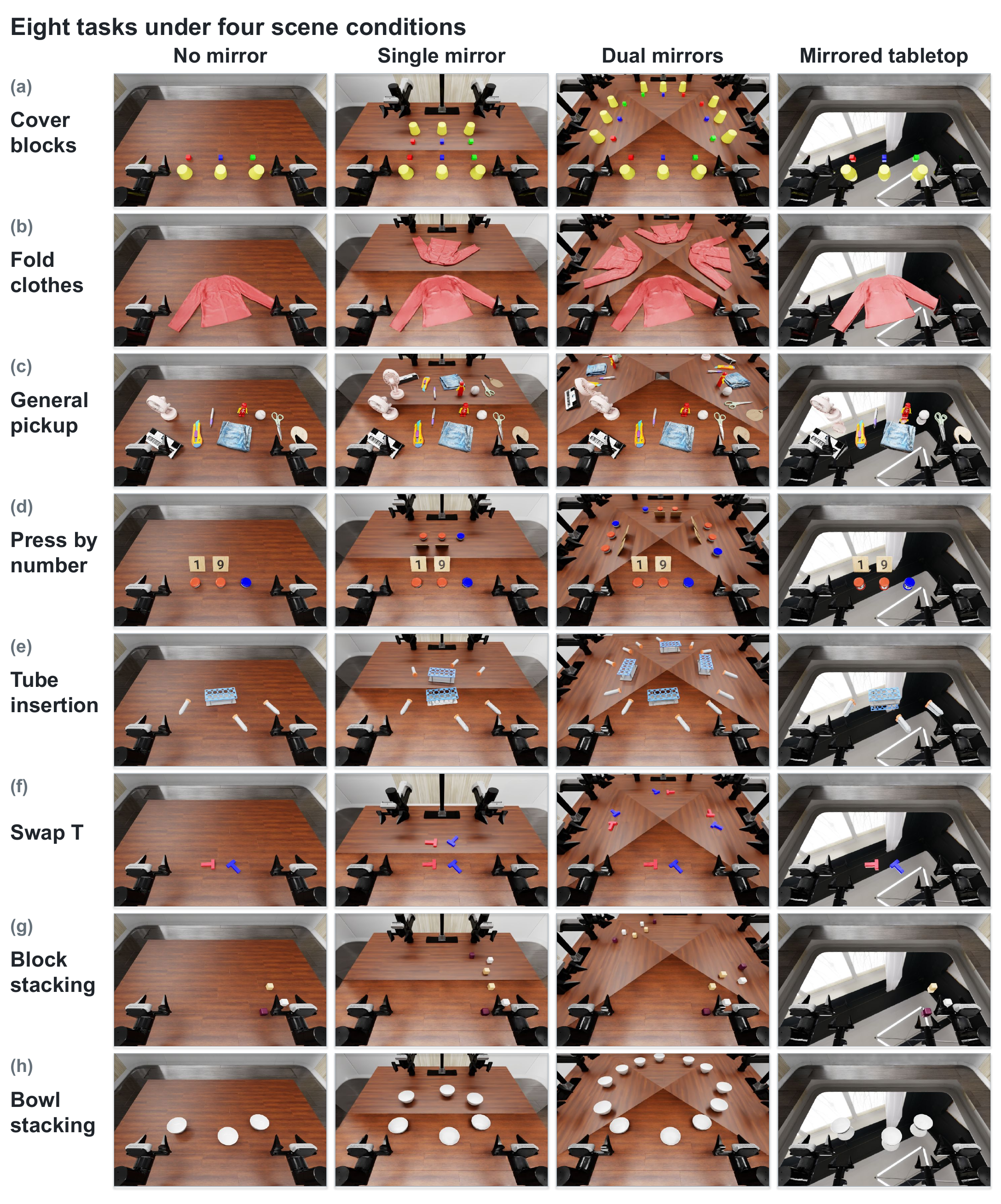}
    \caption{
    The eight common evaluation tasks under four
    scene conditions: no mirror, a single mirror, dual
    mirrors, and a mirrored tabletop.
    Each row shows initial head-camera views from WALL-OSS
    at layout 0 and policy seed 0, with matched robot and
    object initial states across conditions.
    All images use the same crop.
    }
    \label{supp:fig:eight_tasks_four_conditions}
\end{figure*}

We use the same task set across policies and scene
conditions and retain each task's original instructions,
objectives, and native success criteria.
Figure~\ref{supp:fig:eight_tasks_four_conditions}
shows the initial scenes under all four conditions.

\begin{table}[!htbp]
    \centering
    \caption{The eight tasks in the common evaluation set.}
    \label{supp:tab:task_list}
    \small
    \setlength{\tabcolsep}{6pt}
    \begin{tabular}{@{}ll@{}}
        \toprule
        Task & RoboDojo identifier \\
        \midrule
        Cover blocks    & \texttt{cover\_blocks} \\
        Fold clothes    & \texttt{fold\_clothes} \\
        General pickup  & \texttt{general\_pickup} \\
        Press by number & \texttt{press\_by\_number} \\
        Tube insertion  & \texttt{insert\_tubes} \\
        Swap T          & \texttt{swap\_T} \\
        Block stacking  & \texttt{stack\_blocks} \\
        Bowl stacking   & \texttt{stack\_bowls} \\
        \bottomrule
    \end{tabular}
\end{table}

\subsection{Task-Level Performance}
\label{supp:detailed_results}

\begin{table}[tbp]
    \centering
    \caption{
        WALL-OSS-1.0-Lite task-level results.
        Entries report Score / SR (\%).
        Each task--condition cell contains 15 episodes.
    }
    \label{supp:tab:wall_details}
    \footnotesize
    \setlength{\tabcolsep}{1.5pt}
    \renewcommand{\arraystretch}{1.02}
    \begin{tabular*}{\columnwidth}{@{\extracolsep{\fill}}lrrrr@{}}
        \toprule
        Task & \shortstack{No\\mirror} & \shortstack{Single\\mirror}
             & \shortstack{Dual\\mirrors} & \shortstack{Mirrored\\tabletop} \\
        \midrule
        Cover blocks
        & 100.0 / 100.0 & 0.0 / 0.0
        & 1.0 / 0.0 & 0.0 / 0.0 \\
        Fold clothes
        & 89.3 / 86.7 & 69.3 / 66.7
        & 65.3 / 60.0 & 78.7 / 73.3 \\
        General pickup
        & 73.3 / 73.3 & 33.3 / 33.3
        & 20.0 / 20.0 & 60.0 / 60.0 \\
        Press by number
        & 6.7 / 6.7 & 0.0 / 0.0
        & 0.0 / 0.0 & 0.0 / 0.0 \\
        Tube insertion
        & 36.0 / 20.0 & 0.0 / 0.0
        & 0.0 / 0.0 & 5.3 / 0.0 \\
        Swap T
        & 100.0 / 100.0 & 0.0 / 0.0
        & 0.0 / 0.0 & 0.0 / 0.0 \\
        Block stacking
        & 66.0 / 60.0 & 6.0 / 0.0
        & 6.0 / 0.0 & 7.7 / 6.7 \\
        Bowl stacking
        & 93.3 / 93.3 & 77.3 / 73.3
        & 45.0 / 40.0 & 66.0 / 60.0 \\
        \midrule
        Average
        & 70.6 / 67.5 & 23.3 / 21.7
        & 17.2 / 15.0 & 27.2 / 25.0 \\
        \bottomrule
    \end{tabular*}
\end{table}

\begin{table}[tbp]
    \centering
    \caption{
        VPP2 task-level results.
        Entries report Score / SR (\%).
        Each task--condition cell contains 15 episodes.
    }
    \label{supp:tab:vpp2_details}
    \footnotesize
    \setlength{\tabcolsep}{1.5pt}
    \renewcommand{\arraystretch}{1.02}
    \begin{tabular*}{\columnwidth}{@{\extracolsep{\fill}}lrrrr@{}}
        \toprule
        Task & \shortstack{No\\mirror} & \shortstack{Single\\mirror}
             & \shortstack{Dual\\mirrors} & \shortstack{Mirrored\\tabletop} \\
        \midrule
        Cover blocks
        & 100.0 / 100.0 & 0.0 / 0.0
        & 42.0 / 20.0 & 72.0 / 60.0 \\
        Fold clothes
        & 82.7 / 80.0 & 72.0 / 66.7
        & 100.0 / 100.0 & 74.7 / 73.3 \\
        General pickup
        & 6.7 / 6.7 & 20.0 / 20.0
        & 13.3 / 13.3 & 0.0 / 0.0 \\
        Press by number
        & 93.3 / 93.3 & 60.0 / 60.0
        & 73.3 / 73.3 & 46.7 / 46.7 \\
        Tube insertion
        & 96.0 / 93.3 & 13.3 / 0.0
        & 12.0 / 0.0 & 38.7 / 13.3 \\
        Swap T
        & 6.7 / 6.7 & 0.0 / 0.0
        & 0.0 / 0.0 & 0.0 / 0.0 \\
        Block stacking
        & 100.0 / 100.0 & 46.0 / 40.0
        & 15.7 / 6.7 & 6.0 / 0.0 \\
        Bowl stacking
        & 100.0 / 100.0 & 83.0 / 80.0
        & 70.7 / 66.7 & 36.7 / 26.7 \\
        \midrule
        Average
        & 73.2 / 72.5 & 36.8 / 33.3
        & 40.9 / 35.0 & 34.3 / 27.5 \\
        \bottomrule
    \end{tabular*}
\end{table}

\begin{table}[tbp]
    \centering
    \caption{
        Rex-M1 Preview task-level results.
        Entries report Score / SR (\%).
        Each task--condition cell contains 15 episodes.
    }
    \label{supp:tab:rex_details1}
    \footnotesize
    \setlength{\tabcolsep}{1.5pt}
    \renewcommand{\arraystretch}{1.02}
    \begin{tabular*}{\columnwidth}{@{\extracolsep{\fill}}lrrrr@{}}
        \toprule
        Task & \shortstack{No\\mirror} & \shortstack{Single\\mirror}
             & \shortstack{Dual\\mirrors} & \shortstack{Mirrored\\tabletop} \\
        \midrule
        Cover blocks
        & 100.0 / 100.0 & 0.0 / 0.0
        & 0.0 / 0.0 & 0.0 / 0.0 \\
        Fold clothes
        & 93.3 / 93.3 & 49.3 / 40.0
        & 56.0 / 46.7 & 16.0 / 6.7 \\
        General pickup
        & 46.7 / 46.7 & 53.3 / 53.3
        & 26.7 / 26.7 & 6.7 / 6.7 \\
        Press by number
        & 100.0 / 100.0 & 0.0 / 0.0
        & 0.0 / 0.0 & 0.0 / 0.0 \\
        Tube insertion
        & 70.7 / 66.7 & 0.0 / 0.0
        & 1.3 / 0.0 & 0.0 / 0.0 \\
        Swap T
        & 20.0 / 20.0 & 0.0 / 0.0
        & 0.0 / 0.0 & 0.0 / 0.0 \\
        Block stacking
        & 36.7 / 26.7 & 0.0 / 0.0
        & 3.0 / 0.0 & 0.0 / 0.0 \\
        Bowl stacking
        & 83.0 / 80.0 & 41.3 / 33.3
        & 18.7 / 6.7 & 31.0 / 20.0 \\
        \midrule
        Average
        & 68.8 / 66.7 & 18.0 / 15.8
        & 13.2 / 10.0 & 6.7 / 4.2 \\
        \bottomrule
    \end{tabular*}
\end{table}

\begin{table}[tbp]
    \centering
    \caption{
        Awomo-0.5 task-level results.
        Entries report Score / SR (\%).
        Each task--condition cell contains 15 episodes.
    }
    \label{supp:tab:awomo_details}
    \footnotesize
    \setlength{\tabcolsep}{1.5pt}
    \renewcommand{\arraystretch}{1.02}
    \begin{tabular*}{\columnwidth}{@{\extracolsep{\fill}}lrrrr@{}}
        \toprule
        Task & \shortstack{No\\mirror} & \shortstack{Single\\mirror}
             & \shortstack{Dual\\mirrors} & \shortstack{Mirrored\\tabletop} \\
        \midrule
        Cover blocks
        & 100.0 / 100.0 & 0.0 / 0.0
        & 0.7 / 0.0 & 19.7 / 13.3 \\
        Fold clothes
        & 88.0 / 86.7 & 0.0 / 0.0
        & 0.0 / 0.0 & 0.0 / 0.0 \\
        General pickup
        & 66.7 / 66.7 & 40.0 / 40.0
        & 26.7 / 26.7 & 20.0 / 20.0 \\
        Press by number
        & 26.7 / 26.7 & 0.0 / 0.0
        & 0.0 / 0.0 & 0.0 / 0.0 \\
        Tube insertion
        & 78.7 / 73.3 & 57.3 / 46.7
        & 6.7 / 0.0 & 46.7 / 33.3 \\
        Swap T
        & 0.0 / 0.0 & 0.0 / 0.0
        & 0.0 / 0.0 & 0.0 / 0.0 \\
        Block stacking
        & 70.7 / 66.7 & 2.0 / 0.0
        & 2.0 / 0.0 & 0.0 / 0.0 \\
        Bowl stacking
        & 70.7 / 66.7 & 43.0 / 40.0
        & 27.7 / 26.7 & 17.3 / 13.3 \\
        \midrule
        Average
        & 62.7 / 60.8 & 17.8 / 15.8
        & 8.0 / 6.7 & 13.0 / 10.0 \\
        \bottomrule
    \end{tabular*}
\end{table}

\begin{table}[tbp]
    \centering
    \caption{GPT-6-Astra task-level results. Entries report Score / SR (\%).
        Each task--condition cell contains 5 episodes.}
    \label{supp:tab:astra_details}
    \footnotesize
    \setlength{\tabcolsep}{1.5pt}
    \renewcommand{\arraystretch}{1.02}
    \begin{tabular*}{\columnwidth}{@{\extracolsep{\fill}}lrrrr@{}}
        \toprule
        Task & \shortstack{No\\mirror} & \shortstack{Single\\mirror}
             & \shortstack{Dual\\mirrors} & \shortstack{Mirrored\\tabletop} \\
        \midrule
        Cover blocks & 46.0 / 40.0 & 3.0 / 0.0 & 46.0 / 40.0 & 29.0 / 20.0 \\
        Fold clothes & 100.0 / 100.0 & 100.0 / 100.0 & 52.0 / 40.0 & 68.0 / 60.0 \\
        General pickup & 100.0 / 100.0 & 100.0 / 100.0 & 100.0 / 100.0 & 60.0 / 60.0 \\
        Press by number & 80.0 / 80.0 & 100.0 / 100.0 & 80.0 / 80.0 & 60.0 / 60.0 \\
        Tube insertion & 4.0 / 0.0 & 0.0 / 0.0 & 12.0 / 0.0 & 4.0 / 0.0 \\
        Swap T & 20.0 / 20.0 & 20.0 / 20.0 & 20.0 / 20.0 & 40.0 / 40.0 \\
        Block stacking & 100.0 / 100.0 & 100.0 / 100.0 & 83.0 / 80.0 & 83.0 / 80.0 \\
        Bowl stacking & 80.0 / 80.0 & 83.0 / 80.0 & 63.0 / 60.0 & 66.0 / 60.0 \\
        \midrule
        Average & 66.3 / 65.0 & 63.3 / 62.5 & 57.0 / 52.5 & 51.3 / 47.5 \\
        \bottomrule
    \end{tabular*}
\end{table}

The Astra table combines the accepted task-specific
batches. Cover blocks, fold clothes, press by number,
tube insertion, and bowl stacking use layouts 10--14;
general pickup, Swap T, and block stacking use
layouts 0--4. Conditions are paired within each task.
The fold-clothes and press-by-number entries use the
updated matched batch rather than the earlier runs.
No historical and updated executions are pooled into
additional repetitions.

\section{Contact \mbox{Analysis} and Mirror--Panel Controls}
\label{supp:panel_details}

\subsection{Contact Definitions and Locations}
\label{supp:contact_measurement}
We classify contact with an experimental mirror or
non-reflective panel by the entity involved: the gripper
(palm and fingers), other robot bodies (arm links and
wrist-camera bodies where registered), or a task object.
We exclude robot self-contact, robot--object contact, and
contact with the ordinary tabletop. We report each
category where the contact recorder covers the relevant
entities.

Table~\ref{supp:tab:mirror_contact_locations} separates
gripper and other-body contacts to show where contact
occurs in the eight-task benchmark.
For each category, an episode counts once if it contains
at least one recorded contact of that type. An episode
can contribute to both columns for a given configuration.

\begin{table*}[t]
    \centering
    \caption{Contact locations in the eight-task benchmark.
    Entries give the number of episodes with at least one
    recorded contact of the indicated type, over the total
    number of episodes for that policy and configuration.
    Gripper and other-body counts may overlap.}
    \label{supp:tab:mirror_contact_locations}
    \small
    \setlength{\tabcolsep}{7pt}
    \begin{tabular}{@{}lcccc@{}}
        \toprule
        & \multicolumn{2}{c}{Single mirror}
        & \multicolumn{2}{c}{Dual mirrors} \\
        \cmidrule(lr){2-3}\cmidrule(l){4-5}
        Policy & Gripper & Other robot bodies
               & Gripper & Other robot bodies \\
        \midrule
        WALL-OSS       & 28/120 & 40/120 & 27/120 & 30/120 \\
        VPP2           & 32/120 & 55/120 & 53/120 & 52/120 \\
        Awomo-0.5      & 15/120 & 29/120 & 32/120 & 32/120 \\
        Rex-M1 Preview & 11/120 & 14/120 & 52/120 & 40/120 \\
        GPT-6-Astra    &  9/40  & 20/40  & 30/40  & 33/40  \\
        \bottomrule
    \end{tabular}
\end{table*}

For each policy, we use the same contact records as in
Table~\ref{tab:main_results}. An episode is counted as
having robot--surface contact if either the gripper or
another robot body contacts the surface. Task-object
contacts are tracked separately where available and are
excluded from this indicator.

\subsection{Matched Mirror--Panel Controls}
To examine whether direct robot--surface contact accounts
for the observed outcome differences, we compare mirrors
with geometry-matched non-reflective panels using
WALL-OSS and Awomo-0.5. The panels use a light-colored,
matte granite texture. The black seam in the dual-panel
condition is a visual feature. We apply no
condition-specific physics material, friction, or
restitution overrides.

\begin{figure}[t]
    \centering
    \includegraphics[width=\linewidth]{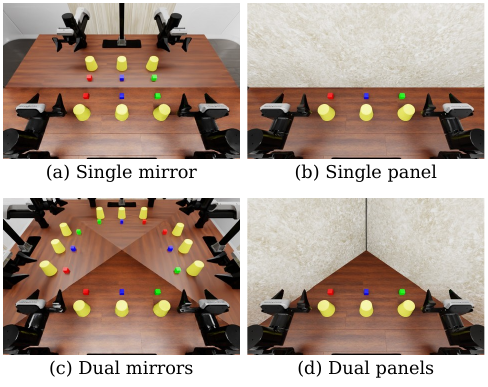}
    \caption{Mirror and non-reflective panel controls in
    \texttt{cover\_blocks}: (a) single mirror, (b) single panel,
    (c) dual mirrors, and (d) dual panels. Each control
    preserves the corresponding mirror geometry and changes
    the surface appearance.}
    \label{fig:panel_ablation}
\end{figure}

Within each policy, the mirror and panel conditions use
matched task layouts and policy seeds. We hold the
checkpoint, task instruction, camera configuration, and
execution budget fixed. Each policy is evaluated on eight
tasks, with five layouts and three seeds per task, giving
120 episodes per condition. The controls cover the
single- and dual-vertical-mirror configurations.
Figure~\ref{fig:panel_ablation} illustrates the setups.

\paragraph{Contact detection for paired controls.}
For WALL-OSS and Awomo-0.5, we query the PhysX GPU
rigid-contact view at each recorded physics substep
($\Delta t=0.004\,\mathrm{s}$). We record contact when at
least one reported contact point between a registered
robot body and the experimental surface has a normal
force greater than $10^{-6}\,\mathrm{N}$.

Let $c_{i,t}^{q}\in\{0,1\}$ indicate robot--surface contact
at physics substep $t$ of episode $i$, where
$q\in\{M,P\}$ denotes the mirror or panel condition.
For dual surfaces, $c_{i,t}^{q}$ is one if contact occurs
with either surface at that substep. The episode-level
indicator is
\begin{equation}
    h_i^{q}=\max_t c_{i,t}^{q}.
\end{equation}

\paragraph{Task-level contact counts.}
Using all matched mirror--panel runs for WALL-OSS and
Awomo-0.5, we compare gripper and other-body contact
event counts across the eight tasks.
For each body category, we count each uninterrupted run
of contacting physics substeps as one event. In the
dual-surface configurations, we take the union of contacts
with either surface before counting events.
Table~\ref{supp:tab:contact_events_by_task} pools both
policies and the single- and dual-surface configurations,
giving 60 mirror episodes and 60 panel episodes per task.
Block stacking has the highest event count for both
body categories on the mirror side.

\begin{table}[t]
    \centering
    \caption{Gripper and other-body contact events by task
    in the matched mirror--panel evaluation. Entries report
    mirror / panel counts. Tasks are ordered by gripper
    contact counts in the mirror conditions.}
    \label{supp:tab:contact_events_by_task}
    \small
    \setlength{\tabcolsep}{4pt}
    \begin{tabular*}{\columnwidth}
        {@{\extracolsep{\fill}}lrr@{}}
        \toprule
        Task & Gripper & Other robot bodies \\
        \midrule
        Block stacking  & 454/264 & 526/385 \\
        Bowl stacking   & 240/107 & 390/218 \\
        Cover blocks    & 239/87  & 355/476 \\
        Swap T          & 230/306 & 188/224 \\
        Tube insertion  &  57/225 &  95/138 \\
        Fold clothes    &  26/166 &   4/16  \\
        General pickup  &   0/1   &   0/4   \\
        Press by number &   0/0   &   0/0   \\
        \bottomrule
    \end{tabular*}
\end{table}

\paragraph{Paired subsets.}
For each matched task, layout, and policy seed, let
$y_i^{B}$, $y_i^{M}$, and $y_i^{P}$ denote native task
success in the no-mirror baseline, mirror, and panel
conditions, respectively. We define
\begin{equation}
\begin{aligned}
    \mathcal F
    &=\{i:y_i^{B}=y_i^{P}=1,\ y_i^{M}=0\},\\
    \mathcal N
    &=\{i:y_i^{B}=1,\ h_i^{M}=h_i^{P}=0\},\\
    \mathcal C
    &=\{i:y_i^{B}=1,\ h_i^{M}=h_i^{P}=1\},\\
    \mathcal F_{00}
    &=\mathcal F\cap\mathcal N.
\end{aligned}
\end{equation}
Here, $\mathcal F$ contains mirror-specific failures.
Among pairs with a successful no-mirror baseline,
$\mathcal N$ contains those with no recorded
robot--surface contact in either condition, and
$\mathcal C$ contains those with recorded contact in
both. Table~\ref{supp:tab:panel_subsets} gives the subset
denominators and task outcomes.

\begin{table*}[t]
    \centering
    \caption{Subset counts for the matched mirror--panel
    comparisons. Within $\mathcal N$, M-only / P-only
    counts pairs with a failure only in the mirror or
    only in the panel condition, respectively. The final
    columns report successes within $\mathcal C$.}
    \label{supp:tab:panel_subsets}
    \small
    \setlength{\tabcolsep}{5pt}
    \begin{tabular}{@{}llcccccc@{}}
        \toprule
        Policy & Config. & $|\mathcal F_{00}|/|\mathcal F|$
        & $|\mathcal N|$ & \shortstack{M-only /\\P-only}
        & $|\mathcal C|$ & \shortstack{M\\success}
        & \shortstack{P\\success} \\
        \midrule
        WALL-OSS  & Single & 7/8   & 41 & 7/2  & 36 & 0/36 & 0/36 \\
                  & Dual   & 11/22 & 35 & 11/1 & 17 & 3/17 & 8/17 \\
        Awomo-0.5 & Single & 9/13  & 47 & 9/5  & 20 & 0/20 & 2/20 \\
                  & Dual   & 20/34 & 40 & 20/2 & 17 & 0/17 & 7/17 \\
        \bottomrule
    \end{tabular}
\end{table*}

\subsection{Contact Persistence}

For each episode, we compute the contact time fraction
as the number of physics substeps with robot--surface
contact divided by the total number of executed physics
substeps. The union over surfaces in $c_{i,t}^{q}$
ensures that simultaneous contact with both panels
contributes only once to this fraction.

Table~\ref{supp:tab:contact_persistence} reports the mean
episode-level fractions within $\mathcal F\cap\mathcal C$,
the mirror-specific failures with recorded robot--surface
contact in both conditions. The ratio divides the
mirror-condition mean by the panel-condition mean.

\begin{table}[t]
    \centering
    \caption{Contact time fractions within
    $\mathcal F\cap\mathcal C$. M and P are mean
    episode-level percentages. Ratios use unrounded
    values. ``--'' denotes an empty subset.}
    \label{supp:tab:contact_persistence}
    \small
    \setlength{\tabcolsep}{4pt}
    \begin{tabular}{@{}llrrrr@{}}
        \toprule
        Policy & Config. & $n$ & M (\%) & P (\%) & M/P \\
        \midrule
        WALL-OSS  & Single & 0 & --    & --    & --   \\
                  & Dual   & 6 & 44.46 & 24.03 & 1.85 \\
        Awomo-0.5 & Single & 2 & 12.18 & 10.93 & 1.11 \\
                  & Dual   & 7 & 32.26 & 4.72  & 6.83 \\
        \bottomrule
    \end{tabular}
\end{table}

\section{Mirrored-Tabletop Grasping Diagnostic}
\label{supp:tabletop}

\subsection{Diagnostic Settings}

We analyze VPP2, Awomo-0.5, WALL-OSS-1.0-Lite,
and GPT-6-Astra on tube insertion, block stacking,
and bottle disposal.
Bottle disposal is an additional diagnostic task
outside the main eight-task set $\mathcal T$.
Each task--condition cell contains 15 episodes
for VPP2, Awomo-0.5, and WALL-OSS, and five for Astra.
Original and mirrored-tabletop conditions are
paired by layout within each policy, with policy
seeds additionally matched for the first three
policies.
This gives 300 executions and 150 matched pairs
across the two conditions.

Task success is taken from the original native
evaluation results.
Behavioral measurements are obtained from saved
commands and physical trajectories.
For VPP2, we also collect reference measurements
by executing the recorded actions in simulation.
The supplementary analysis does not constitute
new policy evaluation.

\subsection{Behavioral Metric Definitions}

\begin{figure*}[t]
    \centering
    \includegraphics[width=\textwidth]{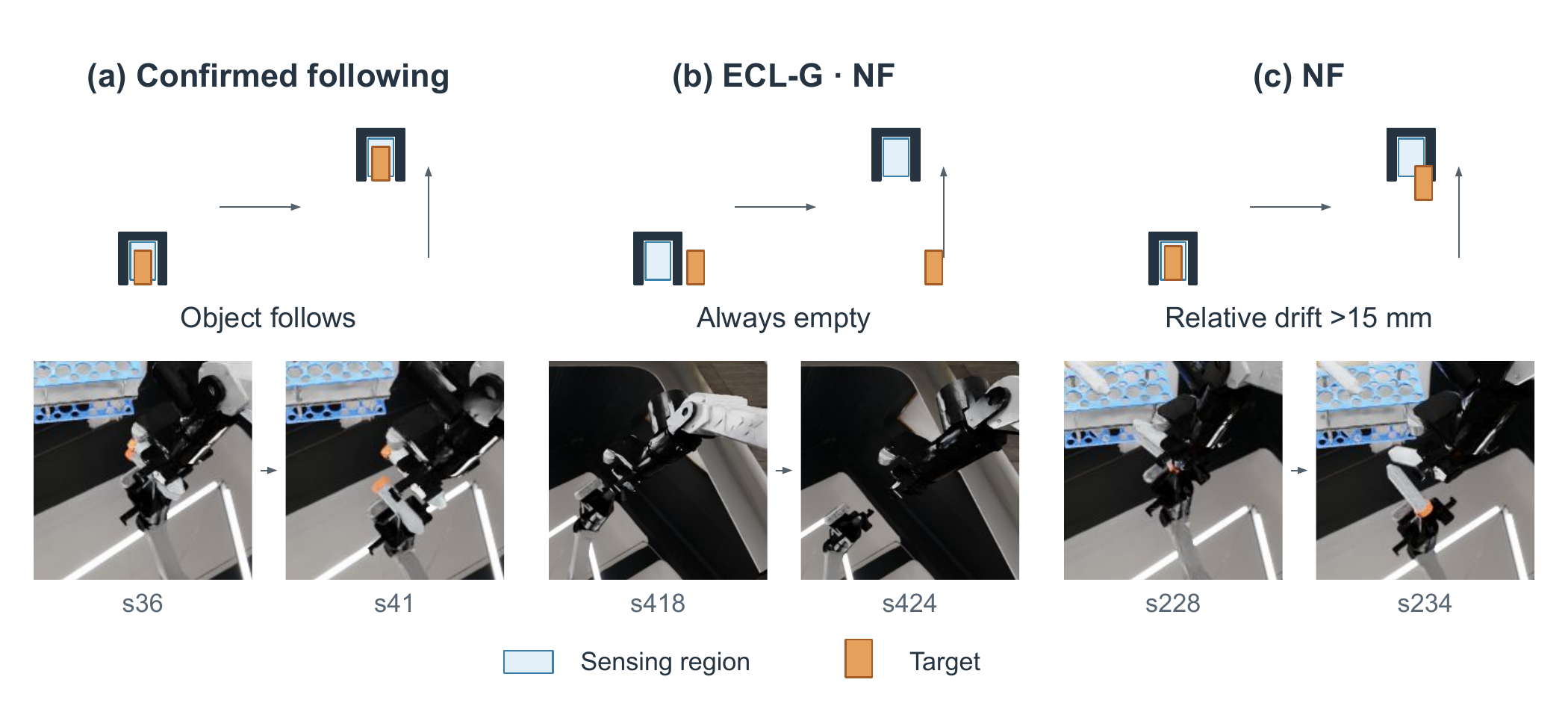}
    \caption{
    Illustrative object-following and empty close-and-lift cases.
    All three examples contain a qualifying TCP lift.
    (a) Confirmed object following.
    (b) Geometric empty close-and-lift (ECL-G), which also satisfies NF.
    (c) Continuous target occupancy with relative palm-frame displacement
    exceeding $15\,\mathrm{mm}$, so object following is not confirmed (NF).
    ECL-G is a subset of NF.
    Camera images show recorded examples; metric labels are determined
    from trajectory measurements.
    }
    \label{supp:fig:ecl_nf_definition}
\end{figure*}

Figure~\ref{supp:fig:ecl_nf_definition} illustrates
object-following and geometric empty close-and-lift
cases under the definitions below.

\begin{figure*}[t]
    \centering
    \includegraphics[width=\textwidth]
        {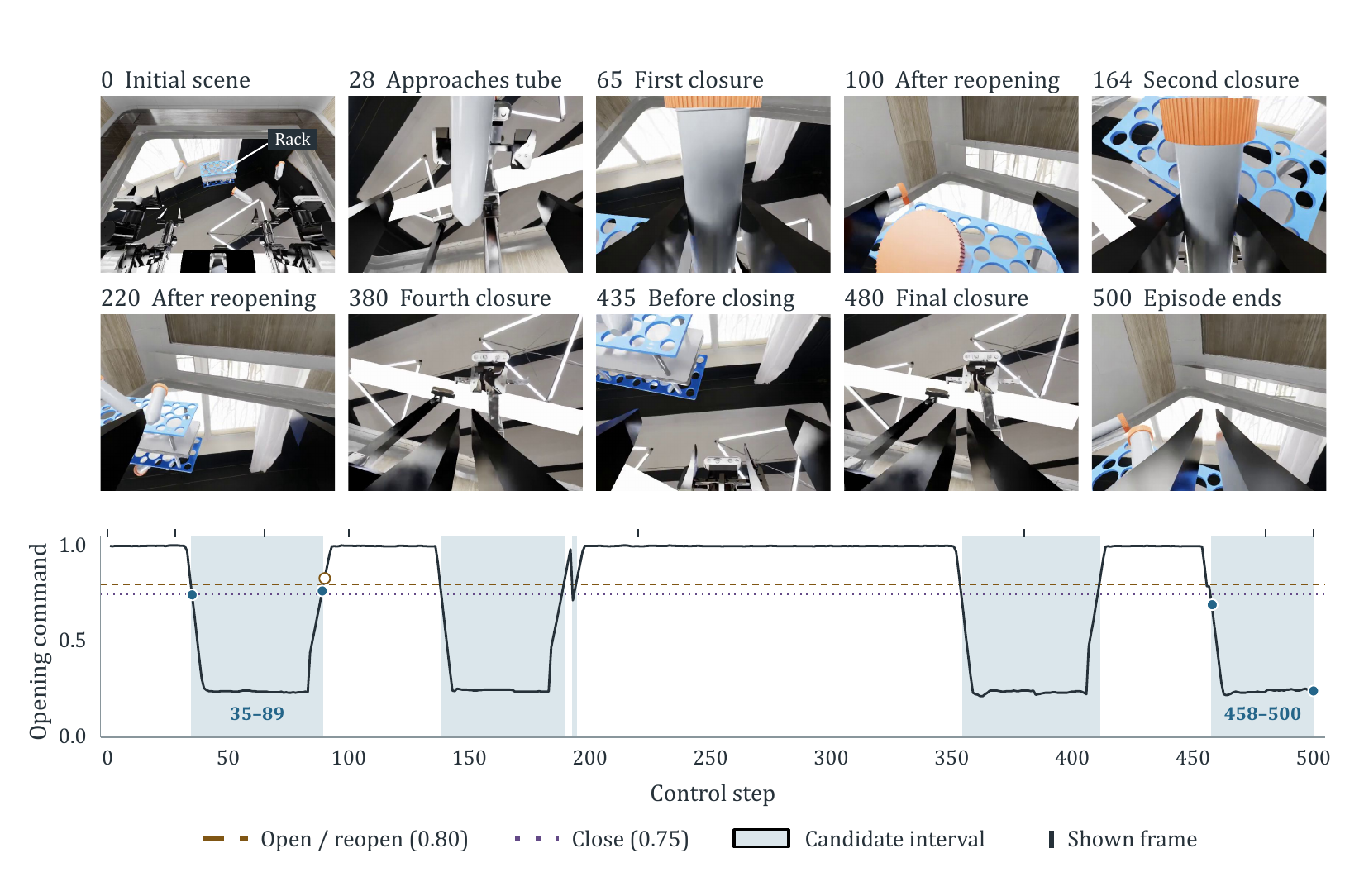}
    \caption{Candidate close intervals in a VPP2 tube-insertion
    episode on the mirrored tabletop. The initial head-camera view
    shows the rack, followed by right-wrist views in chronological
    order. Short ticks above the opening-command curve mark the
    displayed frames, and shading marks all five candidate intervals for the right gripper. The first interval covers
    steps 35--89 and excludes the reopening sample at step 90.
    The final interval covers steps 458--500 and includes the
    last sample because the command stays below the reopening
    threshold through termination. The episode ended without
    completing the task.}
    \label{supp:fig:candidate_close_intervals}
\end{figure*}

\paragraph{Candidate close intervals.}
We identify candidate intervals separately for each arm
using the executed, normalized gripper-opening commands.
After the opening command reaches at least $0.80$,
the first sample at or below $0.75$ starts a candidate
interval. The interval ends at the last sample before
the command returns to at least $0.80$, or at the final
sample of the episode if the command never reaches
the reopening threshold. Figure~\ref{supp:fig:candidate_close_intervals}
shows how these rules identify five candidate intervals
in a recorded tube-insertion episode.

\paragraph{Qualifying lift.}
Within a candidate interval, we search for a TCP
height increase of at least $20\,\mathrm{mm}$
relative to a preceding running minimum.
The height increase must persist for three
consecutive recorded control samples.
This criterion describes TCP motion; it does not
require that an object is lifted.

Let $I=[s,e]$ denote a candidate close interval,
and let $z_t$ be the TCP height at recorded sample $t$.
Define the running minimum

\begin{equation}
    m_t = \min_{u\in[s,t]} z_u.
\end{equation}

We test consecutive three-sample windows within $I$.
For each window starting at $t$, we use the same running
minimum $m_t$ as the reference for all three samples:
\begin{equation}
\begin{aligned}
    L(I)=\mathbf{1}\!\Bigl[
        &\exists\,t\in\{s,\ldots,e-2\}:\\[-2pt]
        &\min_{r\in\{0,1,2\}}(z_{t+r}-m_t)
        \geq 0.02\,\mathrm{m}
    \Bigr].
\end{aligned}
\end{equation}
If $I$ contains fewer than three samples, $L(I)=0$.

\paragraph{Geometric empty close-and-lift (ECL-G).}
We reconstruct an oriented sensing box from the
actual recorded palm and finger poses.
Its lateral extent lies between the inner finger
faces, while its longitudinal and vertical extents
use the shared span of the finger collision meshes.
The region includes a $1\,\mathrm{mm}$ boundary
tolerance.

A candidate is labelled ECL-G if it contains a
qualifying lift and no target collision envelope
intersects the sensing box at any sample from
candidate onset through lift confirmation.
Earlier occupancy prevents an empty label,
even if the object subsequently leaves the region.

Let $O_u$ indicate whether any target collision
envelope intersects the gripper sensing region
at sample $u$. If $L(I)=1$, let $t^\star$ be the first
sample of the earliest qualifying three-sample window,
and let $c=t^\star+2$ be its confirmation sample.
We define
\begin{equation}
    \mathrm{ECL\mbox{-}G}(I)=
    \begin{cases}
        \mathbf{1}\!\left[
            O_u=0\ \forall\,u\in\{s,\ldots,c\}
        \right], & L(I)=1,\\
        0, & L(I)=0.
    \end{cases}
\end{equation}

\paragraph{No lift (NL).}
A candidate is labelled NL when it contains no
qualifying lift before reopening or termination:
$\mathrm{NL}(I)=1-L(I)$.
At episode level, NL indicates that at least one
candidate lacks such a lift.

\paragraph{No confirmed object following (NF).}
For a candidate with a qualifying lift, we test for
object following from the earliest sample attaining
the running minimum $m_{t^\star}$ through the
confirmation sample $c$.
We confirm following if the same target occupies the
gripper sensing region at every sample in this window,
its final height is at least $10\,\mathrm{mm}$ above
its height at the start of the window, and its maximum
Euclidean position deviation from its starting
position, measured in the palm frame, is at most
$15\,\mathrm{mm}$.
We label the candidate NF if it contains a qualifying
lift and no target satisfies all three conditions.

\paragraph{Episode-level aggregation.}
For each metric, we count an episode once if either arm
has at least one candidate interval with that label.
Rates are computed over all evaluated episodes in the
corresponding condition, including successful episodes
and episodes with no candidate interval. An episode can
contribute to multiple metrics when its candidate
intervals receive different labels. Every ECL-G interval
also satisfies NF, whereas NL and NF cannot occur in the
same interval.

\subsection{Task-Level Behavioral Results}

Table~\ref{supp:tab:tabletop_behaviors} reports native
task success, ECL-G, NF, and the auxiliary NL
statistic by task for all four policies.
Each task--condition cell contains 15 episodes for
VPP2, Awomo-0.5, and WALL-OSS, and five for Astra.
Native success comes from the original policy executions;
behavioral measurements use the sources described above.

\begin{table}[!htbp]
    \centering
    \caption{Task-level tabletop diagnostic counts (Original/Mirror).
    Sample sizes are per task and condition.
    Behavioral columns count episodes with at least one event and may overlap.}
    \label{supp:tab:tabletop_behaviors}
    \footnotesize
    \setlength{\tabcolsep}{3pt}
    \renewcommand{\arraystretch}{1.08}
    \begin{tabular*}{\columnwidth}{@{\extracolsep{\fill}}lcccc@{}}
        \toprule
        Task & Success & ECL-G & NF & NL \\
        \midrule
        \multicolumn{5}{@{}l}{\textit{VPP2} ($n=15$)} \\
        Tube insertion & 14/2 & 0/15 & 0/15 & 0/1 \\
        Block stacking & 15/0 & 0/13 & 0/15 & 0/6 \\
        Bottle disposal & 15/0 & 0/15 & 3/15 & 12/12 \\
        \midrule
        \multicolumn{5}{@{}l}{\textit{Awomo-0.5} ($n=15$)} \\
        Tube insertion & 11/5 & 2/4 & 2/6 & 0/1 \\
        Block stacking & 10/0 & 2/0 & 6/2 & 1/12 \\
        Bottle disposal & 15/7 & 0/4 & 0/12 & 15/15 \\
        \midrule
        \multicolumn{5}{@{}l}{\textit{WALL-OSS} ($n=15$)} \\
        Tube insertion & 3/0 & 4/4 & 4/7 & 0/0 \\
        Block stacking & 9/1 & 7/14 & 8/14 & 4/10 \\
        Bottle disposal & 15/12 & 0/0 & 0/2 & 15/14 \\
        \midrule
        \multicolumn{5}{@{}l}{\textit{Astra} ($n=5$)} \\
        Tube insertion & 0/0 & 0/2 & 4/5 & 3/2 \\
        Block stacking & 5/4 & 0/0 & 1/1 & 2/3 \\
        Bottle disposal & 0/1 & 2/2 & 4/5 & 3/2 \\
        \bottomrule
    \end{tabular*}
\end{table}

ECL-G increases across all three VPP2 tasks,
but the pattern is not uniform across policies
or tasks.
For Awomo-0.5, block-stacking detections decrease
from 2/15 to 0/15 despite lower task success.
For WALL-OSS, tube-insertion detections remain
at 4/15, and bottle disposal has no ECL-G
detections in either condition.

NL also coexists with successful manipulation.
For example, all 15 original-tabletop bottle-disposal
episodes succeed for Awomo-0.5 and WALL-OSS,
while every episode contains at least one NL
candidate.

\section{Push-T Diagnostic}
\label{supp:push_t}

\subsection{Scene Construction and Separation Criterion}

We construct three near-mirror configurations in
which a physical T-shaped object and its reflection
form cross-like, $\Pi$-like, or I-like composite
contours in the head-camera view.
The object pose differs across contour configurations.
Consequently, differences between contours may reflect
both visual appearance and physical accessibility.

We distinguish pushing progress from sustained
object--reflection separation.
Let $\mathbf p_{xy}(t)$ denote the object's planar
position at control step $t$.
The displacement is
\begin{equation}
    d(t)=
    \left\|\mathbf p_{xy}(t)-\mathbf p_{xy}(0)\right\|_2.
\end{equation}
An episode reaches the motion threshold if
$\max_t d(t)\geq10\,\mathrm{mm}$.
Maximum displacement is computed over the complete
execution, while steps to $10\,\mathrm{mm}$ are
summarized only over threshold-reaching episodes.

Separation requires a stricter conjunction of
conditions:
\begin{enumerate}
    \item planar displacement of at least
    $10\,\mathrm{mm}$;
    \item a gap of at least five pixels between the
    physical and reflected silhouettes in the fixed
    head-camera view;
    \item both silhouettes remaining in view, with
    at least 50\% red-pixel visibility and at least
    12 visible red pixels each; and
    \item continued table support, with object height
    within $3\,\mathrm{mm}$ of its initial resting
    height and tilt no greater than $15^\circ$.
\end{enumerate}
These conditions must hold continuously for at least
$0.5\,\mathrm{s}$ of simulation time.
The silhouette gap is computed from geometric
projections; visual patches are not subtracted
from the scoring silhouettes.

An episode terminates upon sustained valid separation
or after 1,000 control steps.
Object lifting is recorded but is not an independent
termination condition.
Neither reaching $10\,\mathrm{mm}$ nor satisfying
the separation criterion constitutes success on the
native Push-T alignment objective.

\subsection{Multiple Initial Layouts}

To examine sensitivity to placement, VPP2 and
WALL-OSS-1.0-Lite are evaluated on five matched
layouts per contour, with three policy seeds
per layout.
The layouts apply predetermined lateral translations
of $0$, $0.05$, $0.10$, $0.15$, and $0.20\,\mathrm{m}$
to the reference arrangement. Each policy contributes 45 episodes.

\begin{table}[tbp]
    \centering
    \caption{
        Sustained separation success across five
        lateral placements.
        Each cell contains three policy seeds.
        All collisions remain enabled and no patch
        is applied.
        The final row pools the five layouts.
    }
    \label{supp:tab:push_t_layouts}
    \footnotesize
    \setlength{\tabcolsep}{2pt}
    \renewcommand{\arraystretch}{1.12}
    \begin{tabular}{@{}lcccccc@{}}
        \toprule
        & \multicolumn{3}{c}{VPP2}
        & \multicolumn{3}{c}{\shortstack{WALL-OSS-\\1.0-Lite}} \\
        \cmidrule(lr){2-4}
        \cmidrule(l){5-7}
        \shortstack{Translation\\(m)}
        & Cross-like & $\Pi$-like & I-like
        & Cross-like & $\Pi$-like & I-like \\
        \midrule
        0.00 & 0/3 & 0/3 & 3/3 & 0/3 & 3/3 & 3/3 \\
        0.05 & 0/3 & 0/3 & 3/3 & 0/3 & 2/3 & 3/3 \\
        0.10 & 0/3 & 0/3 & 1/3 & 0/3 & 1/3 & 3/3 \\
        0.15 & 0/3 & 0/3 & 1/3 & 0/3 & 0/3 & 3/3 \\
        0.20 & 0/3 & 0/3 & 0/3 & 0/3 & 0/3 & 0/3 \\
        \midrule
        All layouts
        & 0/15 & 0/15 & 8/15
        & 0/15 & 6/15 & 12/15 \\
        \bottomrule
    \end{tabular}
\end{table}

Table~\ref{supp:tab:push_t_layouts} complements the
displacement-based results in the main paper.
For cross-like configurations, 8/15 VPP2 episodes
and 13/15 WALL episodes reach $10\,\mathrm{mm}$,
yet neither policy achieves sustained valid
separation.
For I-like configurations, the corresponding counts
are 14/15 and 15/15 for reaching $10\,\mathrm{mm}$,
compared with 8/15 and 12/15 for separation.
Thus measurable object motion does not necessarily
satisfy the full separation criterion.

We also evaluate GPT-6-Astra at the same five lateral
offsets, with one execution per offset and contour.
Astra achieves sustained separation in 5/5 executions
for each of the cross-like, $\Pi$-like, and I-like
configurations (15/15 overall). This result uses
the sustained separation criterion defined above.

\begin{figure*}[tp]
    \centering
    \includegraphics[width=\textwidth]
        {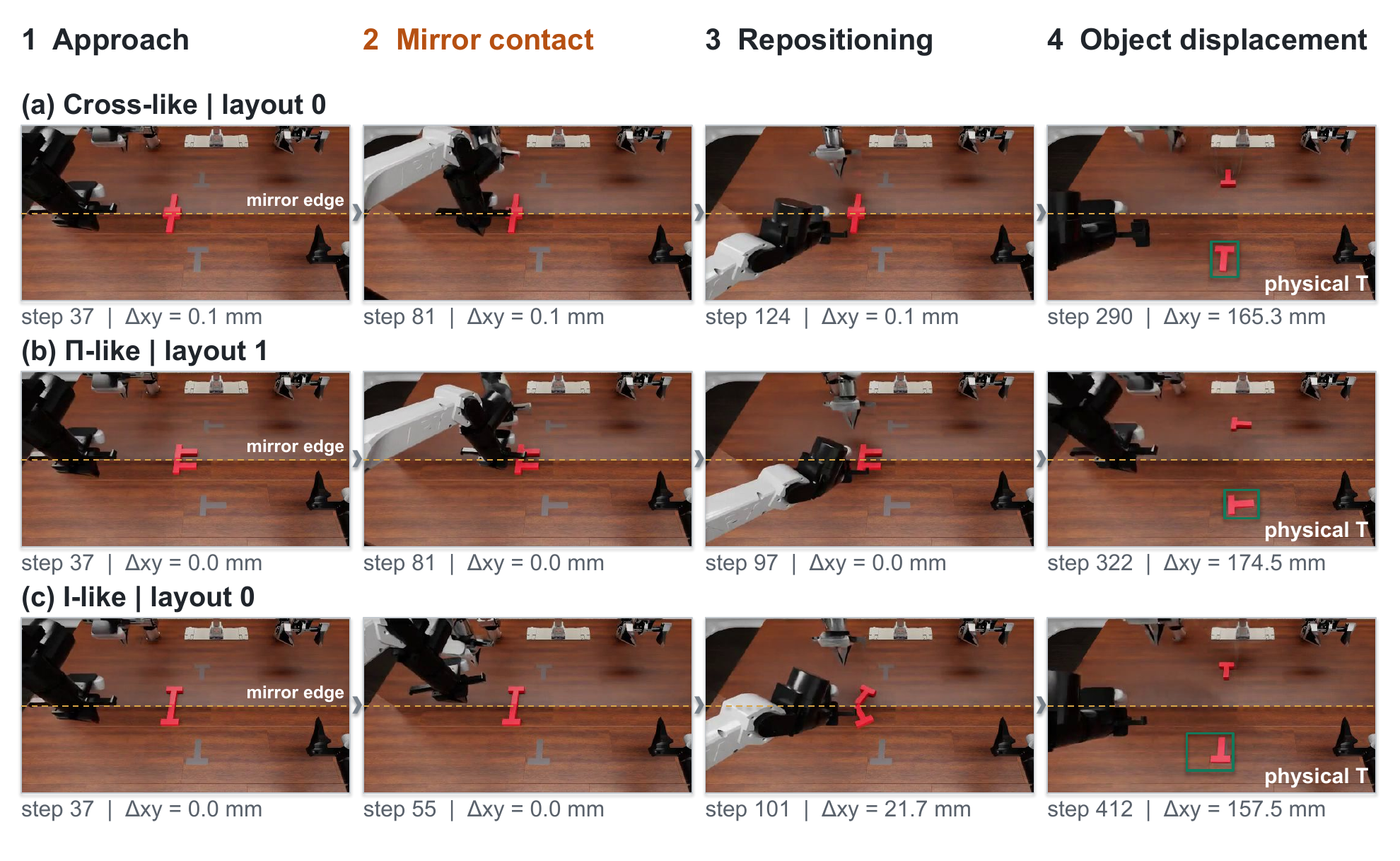}
    \caption{Astra trajectories after mirror contact in
    three Push-T configurations. Each row shows a selected
    episode from approach through contact, repositioning,
    and object displacement. The annotations give the
    control step and the planar displacement of the
    physical object from its initial position.}
    \label{fig:astra_contact_adjustment}
\end{figure*}

Figure~\ref{fig:astra_contact_adjustment} shows one Astra
episode for each Push-T contour. In all three examples,
the robot repositions its end effector after mirror contact
and subsequently displaces the physical T-shaped object.

\subsection{Visual Patch and Collision Controls}

We further evaluate VPP2 at one fixed initial state per
contour using five matched policy seeds.
The visual intervention adds a
$70\times18\,\mathrm{mm}$ matte patch at the
object--reflection junction.
The patch has no collider.

We compare the unmodified mirror and patched mirror
under two collision settings:
robot--mirror collisions enabled, and
robot--mirror collisions selectively filtered.
Object--mirror collisions remain enabled in both.
Within each contour, the checkpoint, task instruction,
camera configuration, and initial state are held
fixed across the compared conditions.
The experiment comprises
$3\times5\times2\times2=60$ executions.

\begin{table*}[t]
    \centering
    \caption{
        VPP2 visual-patch and collision-control results.
        Each cell contains five matched policy seeds
        at one fixed initial state per contour.
        Filtered disables only robot--mirror collisions.
        Control steps are mean $\pm$ sample standard
        deviation across all five episodes, including
        unsuccessful episodes that exhaust the
        1,000-step budget.
    }
    \label{supp:tab:push_t_contact_patch}
    \small
    \setlength{\tabcolsep}{6pt}
    \renewcommand{\arraystretch}{1.12}
    \begin{tabular}{@{}llcccc@{}}
        \toprule
        & & \multicolumn{2}{c}{Separation success}
          & \multicolumn{2}{c}{Control steps} \\
        \cmidrule(lr){3-4}
        \cmidrule(l){5-6}
        Contour & Robot--mirror collisions
        & No patch & Patch & No patch & Patch \\
        \midrule
        Cross-like & Enabled
        & 0/5 & 0/5
        & $1000.0 \pm 0.0$ & $1000.0 \pm 0.0$ \\
        & Filtered
        & 5/5 & 5/5
        & $260.8 \pm 57.5$ & $219.4 \pm 118.3$ \\
        \addlinespace
        $\Pi$-like & Enabled
        & 0/5 & 0/5
        & $1000.0 \pm 0.0$ & $1000.0 \pm 0.0$ \\
        & Filtered
        & 5/5 & 5/5
        & $333.8 \pm 133.2$ & $258.2 \pm 99.9$ \\
        \addlinespace
        I-like & Enabled
        & 5/5 & 3/5
        & $428.6 \pm 51.6$ & $563.4 \pm 403.5$ \\
        & Filtered
        & 5/5 & 5/5
        & $244.0 \pm 94.0$ & $188.6 \pm 100.1$ \\
        \bottomrule
    \end{tabular}
\end{table*}

Table~\ref{supp:tab:push_t_contact_patch} shows that
the patch does not improve separation success when
robot--mirror collisions are enabled.
Cross-like and $\Pi$-like configurations remain at
0/5 under both visual conditions, while the I-like
configuration changes from 5/5 without the patch
to 3/5 with it.

The change from 0/5 to 5/5 in the cross-like and $\Pi$-like
configurations suggests that the physical mirror constrained
VPP2's access to effective pushing motions. One possible
explanation is that VPP2 continued to generate motions
that were obstructed by the mirror. Filtering robot--mirror
collisions allowed such motions to proceed, making the
physical target easier to reach. The filtered results
therefore characterize performance with this physical
constraint removed. The reduction in control steps with
the patch may reflect changes in contour continuity,
local reflected content, or visual salience.

\section{Target-Occlusion Diagnostic}
\label{supp:occlusion}

\subsection{Instruction Conditions}

\begin{figure}[htbp]
    \centering
    \includegraphics[width=\linewidth]
        {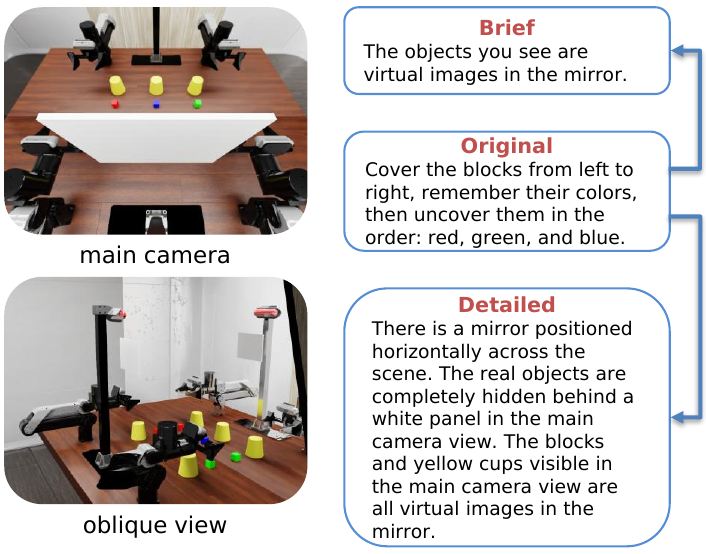}
    \caption{
        Target-occlusion setup and instruction variants
        for \texttt{cover\_blocks}.
        An elevated panel occludes the physical targets
        from the main camera while preserving their
        mirror reflections.
        The oblique view illustrates the physical
        arrangement; it is not an additional policy
        camera.
        Brief and Detailed add the corresponding
        prefixes to the original task instruction.
    }
    \label{fig:floating_panel}
\end{figure}

We compare three instruction conditions. Original uses
the unmodified task instruction. Brief and Detailed
prepend the respective mirror-related prefixes shown
in Fig.~\ref{fig:floating_panel}. Both prefixes refer
to the main-camera view. Some physical targets remain
directly visible to the wrist cameras.

For each policy, we use the same five layouts across
instruction conditions. WALL-OSS and Awomo-0.5 use three
matched policy seeds per layout (15 episodes per condition).
Astra uses one execution per layout (five episodes per
condition). Within each policy, the checkpoint, observation
interface, scene geometry, and execution budget are held fixed.

\paragraph{Target-contact measurement.}
We record contact when the reported normal force between
a robot body and a physical cup or block exceeds
$10^{-6}\,\mathrm{N}$. For each target category, an episode
counts once if it contains at least one qualifying contact.
The contact rate is the fraction of all evaluated episodes
in that policy--instruction condition that meet this criterion.
An episode can contribute to both the cup and block contact
rates.

\subsection{Representative Executions}

\begin{figure*}[tp]
    \centering
    \includegraphics[width=\textwidth]
        {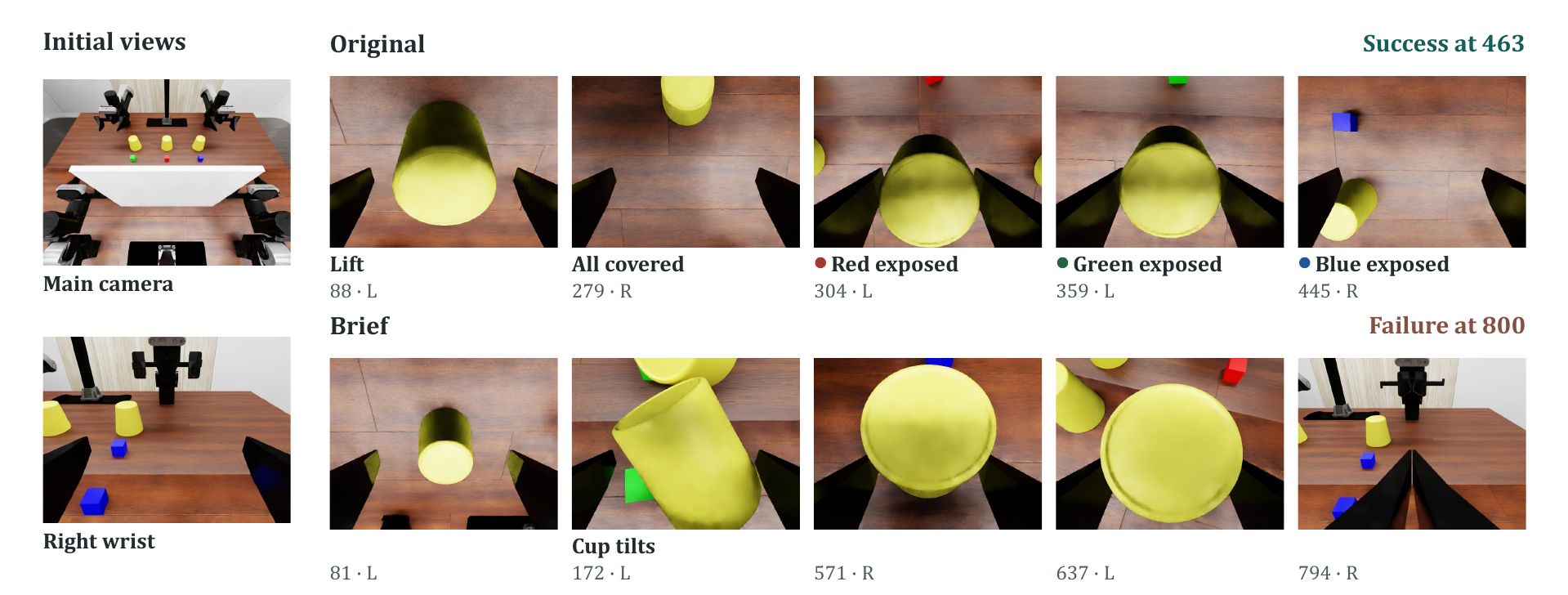}
    \caption{Two Astra executions of \texttt{cover\_blocks}
    under main-camera target occlusion (layout 1).
    The shared initial views show the occluder in the main
    camera and directly visible physical targets in the
    right-wrist camera. The Original execution succeeds at
    control step 463, and the Brief execution ends without
    success at step 800. Numbers below the images give the
    recorded control steps. L and R denote the left and
    right wrist cameras.}
    \label{fig:only_reflection_failure_cases}
\end{figure*}

Figure~\ref{fig:only_reflection_failure_cases} complements
the three-policy results in Table~\ref{tab:reflection_prompt_contact}
with two Astra executions from the same initial layout.
Under Original, Astra covers all three blocks and then
uncovers them in the required order: red, green, and blue.
Under Brief, Astra also moves physical cups, including
one that tilts, but does not complete the required
sequence.

\section{Real-World Experimental Details}
\label{supp:real_world}

\subsection{Robot, Cameras, and Policy Configurations}

We use a Franka FR3 manipulator equipped with a
parallel-jaw gripper.
Visual observations are provided by two Intel
RealSense D435i cameras: a fixed external camera
overlooking the workspace and a wrist-mounted camera.
Both cameras capture RGB images at
$640\times480$ resolution and 30 Hz.
Depth, infrared, and inertial streams are disabled
in the camera configuration.

The policy input includes the two RGB views and
an eight-dimensional proprioceptive state:
seven measured joint positions in radians and
the total gripper width in metres.
The deployed action interface contains seven
absolute joint-position targets and a binary
gripper command, where 0 denotes opening and
1 denotes grasping.
Commands are executed through the ROS 2
joint-impedance control stack.

We use Franka-specific checkpoints for both
$\pi_{0.5}$ and FastWAM.
The $\pi_{0.5}$ deployment uses a LoRA-adapted
multitask checkpoint at training step 31,927,
with a 32-step prediction horizon and a
15 Hz action grid.
FastWAM uses the multitask checkpoint at
training step 30,000, with a 64-step prediction
horizon and a 30 Hz action grid.
Each checkpoint is paired with its saved
normalization statistics.

For $\pi_{0.5}$, camera images are processed
at $224\times224$ resolution.
For FastWAM, the client forwards the separate
native-resolution images to the server, which
applies the checkpoint's training-compatible
image preprocessing.

\begin{table}[t]
    \centering
    \caption{
        Documented Franka policy deployment settings.
        Horizon denotes predicted control steps;
        action rate denotes the command-grid rate.
    }
    \label{supp:tab:real_policy_configs}
    \small
    \setlength{\tabcolsep}{5pt}
    \begin{tabular}{@{}lccc@{}}
        \toprule
        Policy & \shortstack{Checkpoint\\step} & Horizon & \shortstack{Action\\rate} \\
        \midrule
        $\pi_{0.5}$ LoRA & 31,927 & 32 & 15 Hz \\
        FastWAM         & 30,000 & 64 & 30 Hz \\
        \bottomrule
    \end{tabular}
\end{table}

\subsection{Tasks and Mirror Placement}

We consider four tabletop manipulation tasks:
produce placement, block stacking, bracket flipping,
and cylinder insertion.
Table~\ref{supp:tab:real_task_instructions} lists
the task instructions recorded in the deployment
documentation.

\begin{table}[t]
    \centering
    \caption{Task instructions for the real-world evaluation.}
    \label{supp:tab:real_task_instructions}
    \small
    \renewcommand{\arraystretch}{1.12}
    \begin{tabularx}{\columnwidth}
        {@{}l@{\hspace{1em}}>{\raggedright\arraybackslash}X@{}}
        \toprule
        Task & Instruction \\
        \midrule
        Produce placement  & Put the produce in the basket. \\
        Block stacking     & Place the red block on top of the blue block. \\
        Bracket flipping   & Turn the white plastic bracket upside down. \\
        Cylinder insertion & Insert the cylindrical object into the circular holder. \\
        \bottomrule
    \end{tabularx}
\end{table}

Each task is evaluated under three scene conditions:
no mirror, an upright mirror, and a mirrored tabletop.
The upright mirror introduces reflected views beside
the physical workspace.
The mirrored tabletop introduces reflections near
the objects and their supporting surface.
Representative configurations and task initial
states are shown in
Fig.~\ref{fig:real_robot_mirror_cvpr}.

For bracket flipping, the original instruction
is retained verbatim.
Following the recorded demonstrations, success
is defined as reaching a stable upright
configuration.

\subsection{Execution and Recording Protocol}

For each task and policy, we evaluate the same five
initial layouts across the three scene conditions, with
two repetitions per layout. This yields 10 trials per
task, policy, and condition. Table~\ref{tab:real_world_results}
reports the corresponding task-level success rates.

Within each policy, we use the same checkpoint, task
instruction, and camera configuration across scene
conditions. Each policy uses its own deployment
interface and action rate.

\paragraph{Success criteria.}
We assess task success using the following criteria:
\begin{itemize}
    \item \textbf{Produce placement:}
    The produce is fully placed inside the basket
    and released by the gripper.

    \item \textbf{Block stacking:}
    The red block is placed on the blue block
    and remains stable for at least $3\,\mathrm{s}$
    after gripper release.

    \item \textbf{Bracket flipping:}
    The bracket is turned from its initial lying
    pose to a stable upright configuration.

    \item \textbf{Cylinder insertion:}
    The cylinder is stably seated in the receiving
    container and released by the gripper.
\end{itemize}

\paragraph{Trial scoring.}
For canceled recordings, we determine task outcomes
from the available execution and termination records.
Policy-related command rejections, controller alarms,
and safety stops count as failures. Such trials remain
in the denominator of 10 trials per task, policy,
and condition.

\end{document}